\documentclass{article} 
\usepackage[accepted]{icml2024}

\usepackage{amsmath,amsfonts,bm}

\def\eqref#1{equation~\ref{#1}}

\def\1{\bm{1}}

\DeclareMathAlphabet{\mathsfit}{\encodingdefault}{\sfdefault}{m}{sl}
\SetMathAlphabet{\mathsfit}{bold}{\encodingdefault}{\sfdefault}{bx}{n}

\usepackage{hyperref}
\usepackage{url}
\usepackage[utf8]{inputenc} 
\usepackage[T1]{fontenc}    
\usepackage{hyperref}       
\usepackage{url}            
\usepackage{booktabs}       
\usepackage{amsfonts}       
\usepackage{amsmath}
\usepackage{nicefrac}       
\usepackage{microtype}      
\usepackage{xcolor}         
\usepackage{graphicx}
\usepackage{xspace}
\usepackage{caption}
\usepackage{subcaption}
\usepackage{multirow}

\usepackage[capitalise]{cleveref}
\Crefname{equation}{Eq.}{Eqs.}
\Crefname{figure}{Fig.}{Figs.}
\Crefname{tabular}{Tab.}{Tabs.}
\Crefname{section}{Sec.}{Secs.}
\Crefname{appsec}{appendix}{appendices}

\usepackage{wrapfig}
\usepackage{xcolor}
\usepackage{bbm}

\makeatletter
\newcommand{\phantomlabel}[2]{
    \protected@write\@auxout{}{
        \string\newlabel{#2}{
            {\@currentlabel#1}{\thepage}
            {\@currentlabel#1}{#2}{}
        }
    }
    \hypertarget{#2}{}
}
\makeatother

\newcommand{\neww}[1]{\textcolor{black}{{#1}}}

\newcommand{\aharatio}{Eureka-ratio\xspace}

\newcommand{\ahaepoch}{Eureka-epoch\xspace}
\newcommand{\aha}{Eureka-moment\xspace}
\newcommand{\ahas}{Eureka-moments\xspace}

\newcommand{\normsoftmax}{NormSoftmax\xspace}

\icmltitlerunning{Eureka-Moments in Transformers: Multi-Step Tasks Reveal Softmax Induced Optimization  Problems}

\begin{document}
\twocolumn[
 \icmltitle{Eureka-Moments in Transformers:\\Multi-Step Tasks Reveal Softmax Induced Optimization  Problems}
\icmlsetsymbol{equal}{*}

\begin{icmlauthorlist}
\icmlauthor{David T. Hoffmann}{lmb,bosch}
\icmlauthor{Simon Schrodi}{lmb}
\icmlauthor{Jelena Bratuli\'c}{lmb}
\icmlauthor{Nadine Behrmann}{amz}
\icmlauthor{Volker Fischer}{bosch}
\icmlauthor{Thomas Brox}{lmb}
\end{icmlauthorlist}

\icmlaffiliation{lmb}{University of Freiburg}
\icmlaffiliation{bosch}{Bosch Center for AI}
\icmlaffiliation{amz}{Amazon (work done while at Bosch)}

\icmlcorrespondingauthor{David T. Hoffmann}{hoffmann@cs.uni-freiburg.de}

\icmlkeywords{Machine Learning, ICML, sudden convergence, in-context learning, grokking, transformer, Softmax, attention, temperature, gradient, phase transition, abrupt learning, rapid improvement, eureka-moment, eureka moment, multi-step task, multi-step decision task, two-step task}

\vskip 0.3in
]

\newcommand{\fix}{\marginpar{FIX}}

\printAffiliationsAndNotice{} 

\begin{abstract}
In this work, we study rapid improvements of the training loss in transformers when being confronted with multi-step decision tasks. 
We found that transformers struggle to learn the intermediate task and both training and validation loss saturate for hundreds of epochs.
When transformers finally learn the intermediate task, they do this rapidly and unexpectedly. We call these abrupt improvements \emph{\ahas}, since the transformer appears to suddenly learn a previously incomprehensible concept. 
\neww{We designed synthetic tasks to study the problem in detail, but the leaps in performance can be observed also for language modeling and in-context learning (ICL). We suspect that these abrupt transitions are caused by the multi-step nature of these tasks. Indeed, we find connections and show that ways to improve on the synthetic multi-step tasks can be used to improve the training of language modeling and ICL}.
Using the synthetic data we trace the problem back to the Softmax function in the self-attention block of transformers and show ways to alleviate the problem. These fixes reduce the required number of training steps, lead to higher likelihood to learn the intermediate task, to higher final accuracy and training becomes more robust to hyper-parameters.

\end{abstract}

\section{Introduction}
\begin{figure*}[t]
    \begin{subfigure}[]{0.355\textwidth}
        \includegraphics[width=1\linewidth]{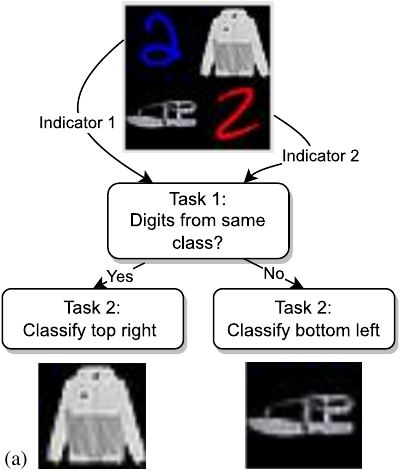}
    \end{subfigure}
    \hspace{0.025\textwidth}
    \begin{subfigure}[]{0.32\textwidth}
        \includegraphics[width=1\linewidth,trim=6 6 6 0,clip]{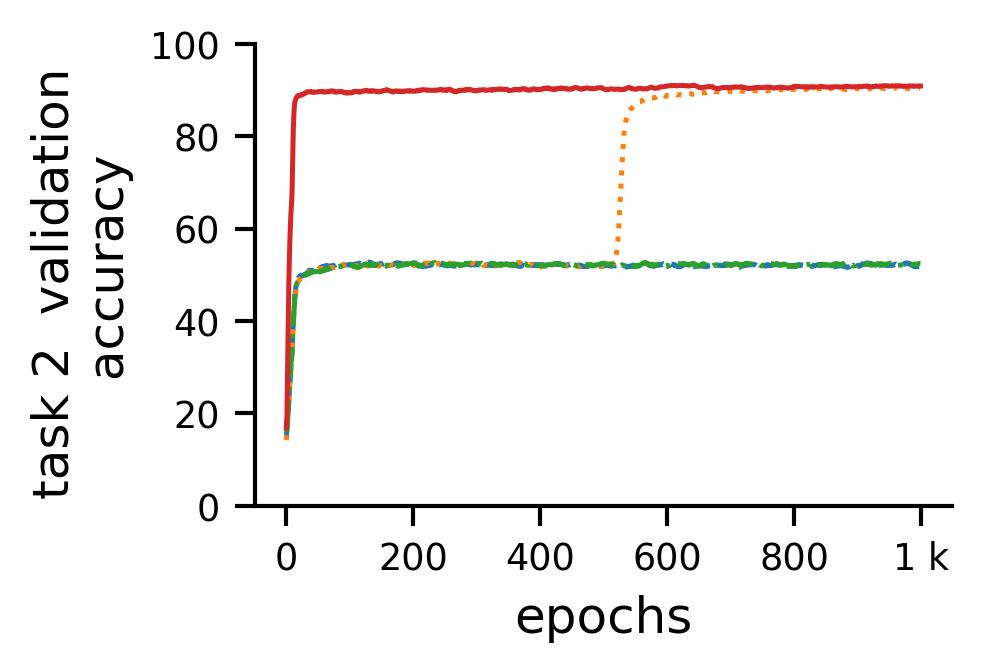}
        \includegraphics[width=1\linewidth,trim=6 6 6 6,clip]{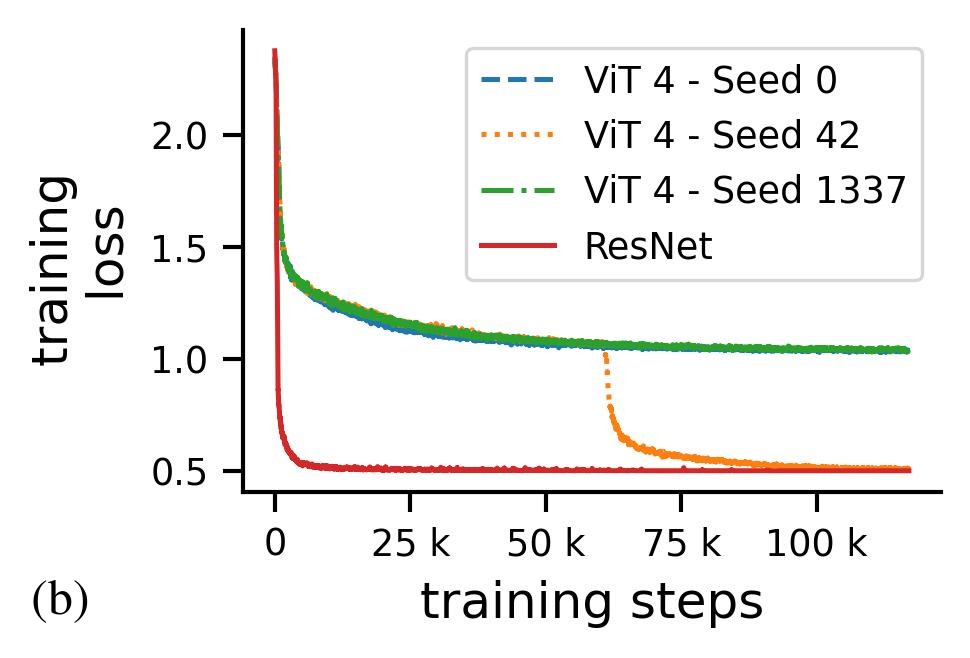}
    \end{subfigure}
    \hspace{0.025\textwidth}
    \begin{subfigure}[]{0.25\linewidth}
        \includegraphics[width=0.8\linewidth,trim=6 6 6 7,clip]{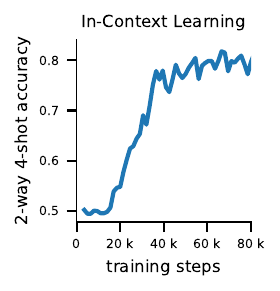}
        \includegraphics[width=0.8\linewidth,trim=6 6 6 6,clip]{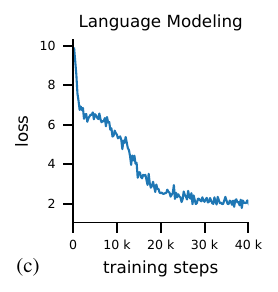}
    \end{subfigure}
    \vspace{-0.5\baselineskip}
    
    \caption{\textbf{Transformers can get stuck during optimization for two-stage tasks.} 
    \textbf{(a)} Describes our 2 step decision task used to study \ahas. \textbf{Task 1} is to compare the two \textbf{indicators} (here digits). If the digits are the same, \textbf{task 2} is to classify the top-right image and bottom left else. Top-right and bottom left are referred to as \textbf{targets}. The location of the correct target is referred to as \textbf{target location}. 
    \textbf{(b)} \textbf{Validation accuracy} and \textbf{training loss} for the task in (a). 2 ViTs (blue and green) fail to converge, while one ViT (yellow) has a \aha. \ahas are characterised by a sudden increase of accuracy and drop of the loss (in contrast to Grokking \citep{grokking}). ResNets are not susceptible to this kind of optimization difficulty.
    \textbf{(c)} \textbf{\ahas on real datasets.} Sharp improvements after initial plateauing can also be observed for GPT-2 ICL, here in the Omniglot ICL task \citep{chan2022data} and language modeling with RoBERTa on Wikipedia. We will show later that our analysis transfers to these tasks (see \cref{fig:real_life_examples_normsoft}).
    }
    \vspace{-1\baselineskip}
    \label{fig:teaser}
    \phantomlabel{a}{fig:teaser_data}
    \phantomlabel{b}{fig:teaser_acc}
    \phantomlabel{c}{fig:real_life_examples}
\end{figure*}

A key quality of any intelligent system is its ability to deal with complex problems that may consist of multiple sub-problems. It should learn to solve these sub-problems even in the absence of direct feedback.
Deep learning has enabled such capabilities to a certain degree. For example, deep classifiers learn the hierarchical feature representations necessary to build a good classifier. 
Reinforcement learning learns object representations required to predict how to receive a sparse reward. 
Language models group tokens to derive their contextual meaning and then predict a new token.
In-context learning (ICL) tasks require to first learn similarities and then associate tokens based on positional information by learning induction heads \citep{olsson2022context}.
While aforementioned examples show great promise, researchers spend a large effort on designing the training process to learn sub-tasks. For instance, reward shaping is common in reinforcement learning, many computer vision works use explicit or implicit intermediate supervision, while for language modelling and ICL a good data arrangement plays an important role \citep{chan2022data}.
For some of these problems, a saturation of the learning process followed by a sudden improvements can be observed, as shown in Fig.~\ref{fig:real_life_examples}.
However, the relation of implicit multi-step learning to saturation of the learning process followed by abrupt improvements in training loss has not been investigated.

But how can we study multi-step learning?
One may tend to study popular tasks in detail, for which many benchmarks and results already exist. 
BERT pretraining \citep{kenton2019bert} and ICL \cite{chan2022data} are candidates that are likely to entail a multi-step task. For BERT the network might first learn word frequencies and might learn to use the context to predict masked tokens in a second step.
Similarly ICL can be understood as multi-step task, where first similarities of tokens have to be learned, followed by learning where to look up the correct label, i.e., learning an induction head. 
For both tasks we observe sudden improvements, similar to those that we study in this work (see Fig.~\ref{fig:real_life_examples}).
Unfortunately, real data prohibits a clean study due to multiple factors:
1) The exact sub-tasks are typically unknown and, hence, hard to study.
2) There are many easy samples that do no require multi-step reasoning,
overlaying the progress on the multi-step task. 
3) The features necessary for the tasks are unknown, i.e.,~we cannot study what the network fails to learn, let alone the reason for it. 
4) Even the number of steps is unknown, thus, we cannot determine if models learn only a subset of the tasks.

As a remedy, we propose to analyze multi-step learning by controlling the data-generating process with synthetic data. This allows us to create clean two-step tasks in a controlled setting that facilitates a detailed study. Specifically, we remove confounding variables, know the number of tasks, and know for each task the relevant features, their location and the total number of steps. 
Thus, it solves the issues above all at once. 
This comes with the assumption that our findings on synthetic data transfer to related observations on real data. Indeed, our understanding how \ahas appear on synthetic tasks and the corresponding way to improve training leads to better training behavior on non synthetic tasks, i.e.,~higher ICL accuracy on Omniglot and earlier \ahas on masked language modeling on Wikipedia (see Fig.~\ref{fig:real_life_examples_normsoft}). 

In each of our synthetic datasets, the answer to the first task $p(z|x)$, which is not explicitly modeled in the loss function, must be found by the model in order to correctly solve the final second task depending on it $p(y|x,z)$. For example, in Fig.~\ref{fig:teaser_data}, the model must first classify the two digits to find out if they are of the same class, which determines where to look for the subsequent FashionMNIST classification task. The loss only provides a training signal for the latter task. Thus, the model must figure this out by itself during training. Formally, such multi-step tasks can be described as $p(y|x,z) \cdot p(z|x)$, i.e.,~the probability of class $y$ given evidence $x$ and the latent variable $z$.

Our study reveals that transformers have difficulties in learning such two-step tasks (Fig.~\ref{fig:teaser_acc}).
After they learned to classify a randomly selected FashionMNIST image $p(y|x,z)$, they  saturate and only after a long time suddenly learn $p(z|x)$, i.e.,~the task to select the right fashion image by comparing the digits.
We call this phenomenon a \aha. Intriguingly, we find that they never learn task 1 within 1000 epochs and stick with the prior $p(z)$ for some random seeds. We later find that the probability of \ahas depends on the difficulty of the task. In contrast to the transformer, a ResNet learns both tasks immediately.  

Our goal is not to add to the old transformer vs. CNN discussion, but we want to investigate this particular problem. What is its cause? Is it due to a too small capacity? Is it the number of heads or the learning rate? Is it the spatial arrangement of task 1 and 2? We found that these factors play only a minor role and finally traced the problem back to the Softmax function in the transformer's attention blocks. We found parts of the gradient's components become very small depending on the Softmax output.
This cannot be fixed by using a larger learning rate, since other components of the gradient are large, but it can be simply fixed by a normalization of the Softmax function.

In summary, the contributions of this analysis paper are: 1) We study multi-step learning without intermediate supervision via a fully controlled data-generating process on synthetic tasks. 2) We discover a new failure mode of transformers.
3) We analyse the mechanisms underlying this failure mode and find that the Softmax function leads to (local) small gradients for the key and value weight matrices, thereby hindering learning. 
4) To validate the role of Softmax, we mitigate the failure mode through targeted interventions.
We show that these interventions lead to significantly faster convergence, higher accuracy, higher robustness to suboptimal hyper-parameters, and higher probability of model convergence, affirming our analysis. 
5) We find related learning behavior in ICL and language modeling and show that our solution transfers to these settings.
The code to reproduce the results and create the datasets is available\footnote{\url{https://github.com/boschresearch/eurekaMoments}}.

\section{Related Works}
\textbf{Emergence and phase transitions.}
Following \citet{steinhardt_ml_systems_will_be_different} \& \citet{wei2022emergent},
\emph{emergence} refers to a qualitative change in a learning system resulting from an increase of model size, training data or training steps, where \emph{phase transitions} are additionally characterized by a sharp change. \ahas are special types of phase transitions.
While recent work showed that sharp emergence may just be an artifact of the choice of a discontinuous metric~\citep{emergancemirage,srivastava2022beyond}, we observe rapid changes also for continuous metrics in our setting.
A connection between phase transitions or emergence and our work may exist, and both may be related to escaping bad energy landscapes as in our work.

\textbf{Unexplained phase transitions.}
Previous works reported observations that may be \ahas, without investigating their cause. 
For instance, rapid improvements happen for in-context-learning (ICL) \citep{olsson2022context}, diffusion model training \cite{zhang2023adding} and BERT training \citep{gupta2020gmat,attentionGuiding,nagatsuka2021pre,anonymous2023,chen2024sudden}.
\citet{attentionGuiding} proposed to bias the attention mechanism towards predefined attention patterns and observed speed-ups in BERT training, while \citet{chen2024sudden} connect the sharp drop to sudden learning of syntactic attention structures.
We also identify the learning of the task-required attention pattern to be the cause of the problem.
In concurrent work \citet{reddy2023mechanistic} studied phase transitions in ICL. Specifically, they show for a small toy-model that slow initial learning is due to a saddle-point, where one path leads to a sub-optimal minimum (i.e.,~random context label) and the other path to the ICL solution. They discover a ``cliff'' in the loss landscape and shallow gradients that lead to the ICL solution. While our setting shows substantial differences, our analysis reveals the same underlying mechanisms, i.e.,~ shallow gradients, which we investigate in more detail.

\textbf{Grokking.}
A similar phenomenon has been discovered on synthetic data \citep{grokking} and was further studied in \citep{omnigrok,mechanistic_grokking,thilak2022the,millidge2022grokking,barak2022hidden,liu2022towards}. Grokking describes the phenomenon of sudden generalization after overfitting, which can be induced by weight decay. 
In contrast to \ahas, the training accuracy already saturates at close to 100\% (overfitting), a long time before the validation accuracy has a sudden leap from chance level to perfect generalization. 
For \ahas, validation and training loss saturate (no overfitting) and the sudden leap occurs for \emph{both simultaneously}.

\textbf{Unstable gradients in transformers.}
The position of the layer-norm (LN) \citep{xiong2020layer} and instabilities in the Adam optimizer in combination with LN induced vanishing gradients \citep{huang2020improving}. Removing the LN \citep{baevski2018adaptive,child2019generating,wang2019learning} or Warmup \citep{baevski2018adaptive,child2019generating,wang2019learning,huang2020improving} resolves this problem, but in our case, Warmup alone does not help.
Others identified the Softmax as one of the problems, showing that both extremes, attention entropy collapse, i.e.,~too centralized attention \citep{att_entropy_collapse,relu_softmax} and a large number of small attention scores, i.e.,~close to maximum entropy \citep{attentionisnotallyouneed,trivial_attention} can lead to small gradients \citep{noci2022signal}.
As a remedy to vanishing gradients caused by entropy collapse \citet{wang2021escaping} proposed to replace the Softmax by periodic functions.
However, before \ahas, the attention distribution is in neither extreme. Instead the attention is allocated to the wrong tokens. 

\textbf{Temperature in Softmax.}
A key operation in the transformer is the scaled dot-product attention.
Large products can lead to attention entropy collapse \citep{att_entropy_collapse,relu_softmax}, which results in very small gradients. 
In contrast, \citet{trivial_attention} observed close to uniform attention over tokens.
They scaled down very low scores further while amplifying larger scores, but this only amplified the problem when important tokens are already ignored.
Instead, \citet{normsoftmax} proposed to normalize the dot product.
Their proposed \textit{\normsoftmax} avoids low variance attention weights and thus avoids the small gradient problem. We found it as the most effective intervention on the Softmax function.
Others proposed to learn the temperature parameter \citep{learn_temp,ali2021xcit}, but this is difficult to optimize.
For very large models the problem becomes more severe.
Models with more than 8B parameters show attention entropy collapse \citep{vit22b}.
They followed \cite{gilmer2023intriguing} and normalized the $QK^T$ with layer norm before the Softmax.

\section{Background}
\label{sec:background}
\textbf{Preliminaries. }
This work investigates the popular dot product attention \citep{transformer}, defined as
\begin{equation}
    \text{Attention}(Q,K,V) = \text{Softmax}\left(\frac{QK^T}{\tau}\right)V,
\end{equation}
where the weight matrices $W_Q$, $W_K$ and $W_V$ map the input $X$ to query $Q$, key $K$, value $V$, and the temperature parameter $\tau$ controls the entropy of the output. A low temperature leads to low entropy, i.e.,~a more ``peaky'' distribution. 
Commonly, $\tau$ is set to $\sqrt{d_k}$, where $d_k$ is the dimensionality of $Q$ and $K$. Thus, $\sqrt{d_k}$ is the standard deviation of $QK^T$ under the independence assumption of rows of $Q$ and $K$ with 0 mean and variance of 1 \citep{transformer}.

\textbf{Softmax attention can cause vanishing gradients.}
Attention entropy collapse, i.e.,~too centralized attention, can cause vanishing gradients \citep{att_entropy_collapse,relu_softmax}, since all entries of the Jacobian of the Softmax will become almost 0 (see .~\ref{sec:vanishing_gradient}).
Similarly, uniform attention can cause vanishing gradients for $W_K$ and $W_Q$ \citep{noci2022signal}.

A remedy to both problems is to control the attention temperature $\tau$.
A larger $\tau$ in the Softmax will dampen differences of $QK^T$ and by that prevent vanishing gradients by low attention entropy.
In contrast, a smaller $\tau$ will amplify differences of $QK^T$ and prevents vanishing gradients caused by uniform attention. 
Choosing the right temperature is difficult and can have a strong influence on what the model learns, how fast it converges etc. As a remedy, we propose to start training with a low temperature and follow a schedule to heat it up to the default value of $\sqrt{d_k}$.
We refer to this approach as \textbf{Heat Treatment} (HT) and it is one of our interventions to test whether the Softmax is indeed the root cause of the optimization difficulties.
This approach has multiple advantages. First, it removes the difficulty of choosing the exact temperature. Second, the network gets optimized for ``more peaky'' attention, but the temperature increases steadily. By that, the network starts with centralized attention but since the next epochs attention will be more uniform than the previous (due to increasing the temperature), it does not run into the issue of low attention entropy. 
Last, the network can focus on most important features early in training and broaden the view over time, attending to other features.

\textbf{\normsoftmax.}
An alternative to tame the attention is \normsoftmax \citep{normsoftmax}, which replaces the expected standard deviation $\sqrt{d_k}$ gets by the empirical standard deviation $\sigma(QK^T)$, for each attention block individually. \normsoftmax can be computed by 
\begin{equation}
    \text{NormSoftmax}(Q,K)=\text{Softmax}\left( \frac{QK^T}{min(\sigma(QK^T), \tau)}\right).
\end{equation}
If $QK^T$ has low standard deviation differences will be amplified. If $\sigma(QK^T)>\tau$, $\tau$ will be used. 

\section{Task Description \& Experimental Conditions}
\label{sec:task_description}
Recent works across diverse fields found sudden abrupt learning behaviors, e.g.,~sudden improvements of RoBERTa \citep{roberta}, rapid emergence of induction heads \citep{olsson2022context}, or the ``sudden convergence phenomenon'' of control net \citep{zhang2023adding}.
We hypothesize that all of these constitute multi-step mechanisms $p(y|x,z) \cdot p(z|x)$, where $p(y|x,z)$ (task 2) is dependent on the result of $p(z|x)$ (task 1), where only the last task is supervised. 
But why does it take so long for transformers to learn such mechanisms and why is the improvement so sudden?

\textbf{Task description.}
We suspect that these training problems are due to the multi-step nature of these tasks. To test this we study such multi-step mechanisms on a simple two-step tasks. Note that many-step tasks would also be possible but are a more complicated study object.
Fig.~\ref{fig:teaser_data} provides a schematic overview for one of our vision tasks:
Task 1 requires the model to indicate where to look at, i.e., top right or bottom left depending on whether the MNIST digits \citep{mnist} in the top left and bottom right match or do not match. Task 2 is a simple classification. Here it is FashionMNIST \citep{fashionmnist} classification.
Note that only task 2 is evaluated and only task 2 gets supervision; akin to, ICL for example, for which supervision is only provided for a the missing token, but not directly for the induction head learning.
By design of our datasets, $40-55\%$ accuracy can be obtained by only solving task 2 and picking the target at random. The range is due to varying difficulty of task 2. Higher accuracies can only be achieved by learning the multi-step structure.

\textbf{Vision dataset creation.}
The visual datasets are based on MNIST \citep{mnist} and Fashion-MNIST \citep{fashionmnist}. 
An example and a schematic of the task is shown in Fig.~\ref{fig:teaser_data}.
The samples are created by sampling 2 random Fashion-MNIST samples and 2 digit samples from the MNIST classes ``1'' and ``2''. We apply a random color to the MNIST samples (red or blue). Next, we compose a new image from the 4 images, putting the 2 MNIST samples on top left and bottom right and the Fashion-MNIST samples on the remaining free quadrants. If the 2 MNIST samples are from the same class, the class of the top-right image is the sample label and bottom-left else. 

\textbf{Reasoning task.} 
Complementary to above vision tasks, we further simplified the multi-step task to an algorithmic task of the form
$    f(a,\;b,\;c,\;d)=\left\{\begin{array}{ll} c, & g(a,\;b) \\
         d, & \textrm{otherwise}\end{array}\right.,
         \label{eq:reason_eq}$
where \begin{math}a,b,c,d\in\{0,\;1,\;...,\;n\}\end{math}, task 1 is \begin{math}g(a,\;b)=\mathbbm{1}[(\textrm{a is even}\land\textrm{b is odd}) \lor (\textrm{a is odd}\land\textrm{b is even})]\end{math} and task 2 is a simple copying task of either $c$ or $d$. More details are provided in \ref{sec:reasoning_task_suppmat}.

\textbf{Metrics.}
We define the \textbf{\aharatio} as the proportion of training runs with \aha across the different random seeds.
To automatically detect \ahas, we set a conservative threshold at a validation accuracy of $70\%$, as this threshold can not be crossed without solving task 1. 
\textbf{Accuracy after \aha}, in the following referred to as accuracy, is computed over all runs that had a \aha. Note that this metric must be jointly considered with the \aharatio, since high accuracy with low \aharatio indicates that optimization typically failed (no \aha) but possibly just one ``lucky'' training run learned to solve the multi-step mechanism.
Finally, \textbf{average \ahaepoch} provides the average epoch at which the \aha happened. It is computed only for runs with \aha and again must be interpreted jointly with the \aharatio.

\textbf{Models and hyper-parameters.}
Following \citet{compactvit}, we use a ViT version specifically designed for small datasets. 
Unless stated otherwise, we train a ViT with 7 layers, 4 heads each with embedding dimension of 64, patch size of 4 and MLP-ratio of 2. Consequently, the default temperature is $\sqrt{d_k}=8$. The ResNet has a comparable parameter count and consists of 9 layers. 
For ViT, ViT+HT and \normsoftmax, we tested 5 different temperature parameters in initial experiments. More details on the training setup are provided in ~\ref{sec:experimental_setup_vision}.
For the reasoning task, we train a transformer on 30\% of the entire set of possible inputs (i.e.,~\begin{math}11^4=14\,641\end{math} input combinations) for 10K epochs over five random seeds. More details on model and training are provided in \ref{sec:reasoning_task_suppmat} .

\section{Understanding \ahas \& Optimization Problems of Transformers}
\begin{figure}[t]
    \centering
    \begin{subfigure}[l]{1.0\linewidth}
        \includegraphics[width=0.935\linewidth]{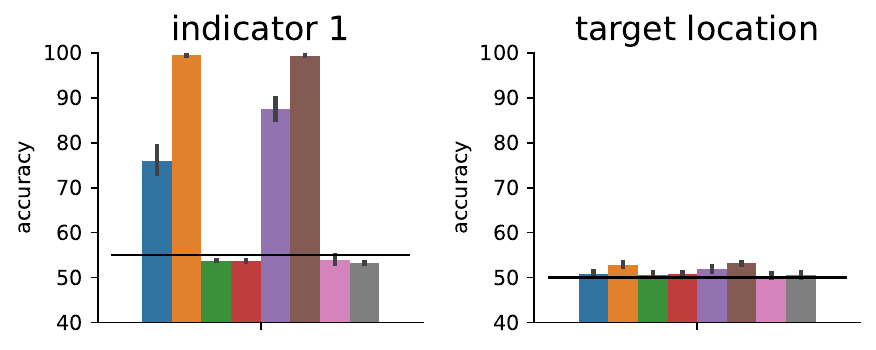}
        \caption{ViT (without \aha)}
    \end{subfigure}
    \begin{subfigure}[l]{1.0\linewidth}
        \includegraphics[width=1\linewidth]{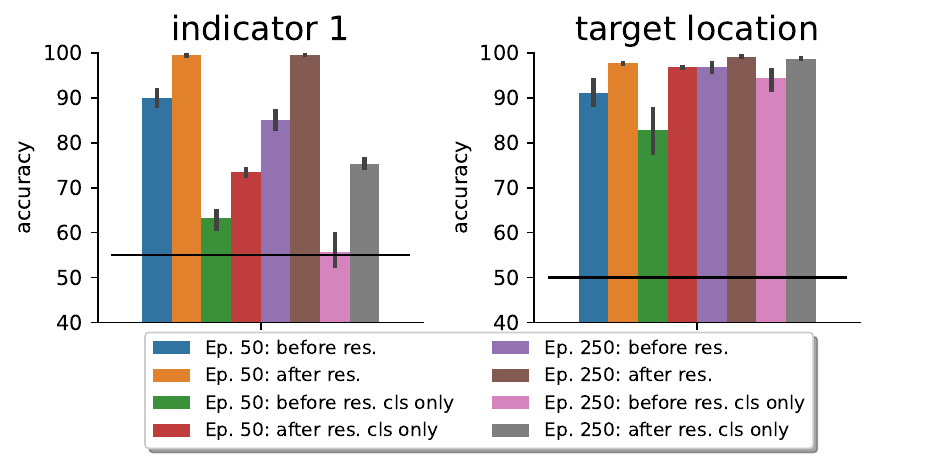}
        \caption{ViT $\tau=\frac{1}{3}$ (with \aha)}
    \end{subfigure}
    \vspace{-0.75\baselineskip}
    \caption{\textbf{What is represented in different parts of the attention block.} Bar plots show linear probe accuracy averaged over heads. Indicator 1 is the top MNIST digit. Both ViT and ViT $\tau=\frac{1}{3}$ extract the indicator class information from the images and it is available in each layer. Information is available before and after the residual connection, therefore it is not entirely ignored by the attention. Differences between ViT and ViT $\tau=\frac{1}{3}$ are visible for CLS token and target location task. Res.~denotes residual layer. Error bars show variance over heads. Results for layer 6 using $Z_i$. Black line indicates chance. Indicator 2 plots are similar. More layers and indicator 2 plots are shown in Fig.~\ref{fig:linprobe_z}. $Q$, $K$, $V$ linear probes in Figs.~\ref{fig:linprobe_q} to \ref{fig:linprobe_v}. A similar analysis using more sensitive information-theoretic probes \citep{voita2020information} can be found in \cref{sec:info_probes}}
    \label{fig:linprobe_vit_failure}
  \phantomlabel{a}{fig:linprobe_vit_failure_failure}
  \phantomlabel{b}{fig:linprobe_vit_failure_success}
  \vspace{-1\baselineskip}
\end{figure}

\begin{figure}[h]
    \begin{subfigure}[c]{1\linewidth}
    \centering
    \includegraphics[width=0.8\linewidth,trim=6 11 0 3,clip]{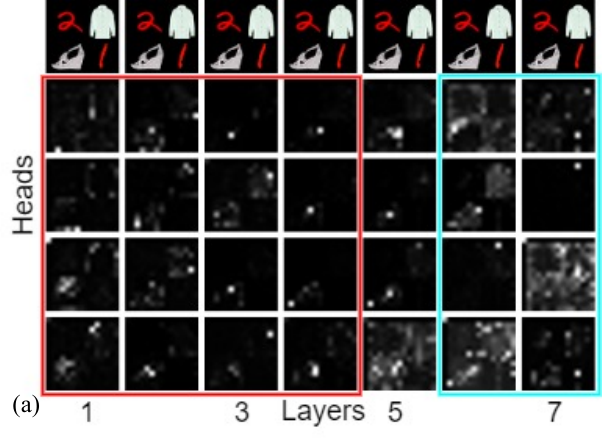}
    \end{subfigure}
    \begin{subfigure}[c]{1\linewidth}
    \centering
    \includegraphics[width=0.8\linewidth,trim=6 11 0 3,clip]{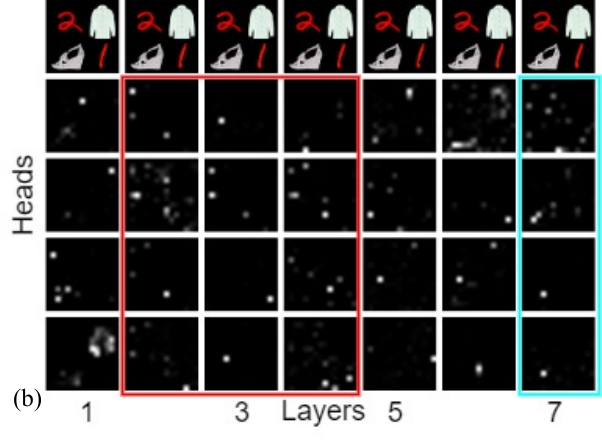}
    \end{subfigure}
    \vspace{-0.75\baselineskip}
     \caption{\textbf{Attention maps after training for:}  \textbf{(a)} \textbf{ViT without \aha}. It fails to compare the 2 digits. First layers explicitly ignore indicators (digits) (highlighted with red). \textbf{(b)}\textbf{ ViT $\tau=\frac{1}{3}$ with \aha} attends indicators in first layers (red) and predominantly attends the correct target (ankle boot) in later layers. Black is no and white is high attention. Maps show the average attention of each query, i.e.,~we average over the key-dimension of the attention map.}
      \phantomlabel{a}{fig:attn_maps_failure}
     \phantomlabel{b}{fig:attn_maps_success}
    \label{fig:attn_maps}
    \vspace{-1\baselineskip}
\end{figure}

Here, we analyze the problem on the dataset described in Fig.~\ref{fig:teaser_data}. In Sec.~\ref{sec:artifical_problem} we provide supporting experiments and finally show indications, that the results can be transferred to real datasets, i.e.,~Wikipedia text completion and ICL.

\subsection{Why Do Transformers Fail to Learn Two-step Tasks?}
To investigate why ViT's training fails, we analyze the learned representations using progress measures.
Note that solving task 1 requires ViT to \textbf{1)} learn to distinguish the indicators, \textbf{2)} carry the information through the layers and \textbf{3)} compare the indicator information to obtain the target location.
We use linear probes on the output of the attention heads, i.e.,~$Z_i =\text{Attention}(Q_i,K_i,V_i)$, for all heads $i$.
However, note that linear separation becomes very likely as feature dimensionality increases. Thus, a high accuracy on the linear probe classification does not imply that the transformer is using that information. It only shows that the information is represented and linearly separable.

\textbf{Does the transformer fail to distinguish the indicators?}
The left bar plots in Fig.~\ref{fig:linprobe_vit_failure_failure} show that the indicators can be linearly separated well (orange, brown) across all layers (Fig.~\ref{fig:linprobe_z}). Nonetheless, ViT fails to learn task 1 (Fig.~\ref{fig:linprobe_vit_failure_failure} right plot). 
Thus, the transformer represents the information to solve task 1 but does not utilize it.

\textbf{Does the transformer filter out information required for task 1?}
Since the loss provides a training signal only for task 2, the transformer may learn to just ignore the indicators.
Task 1 information could be ignored in the attention blocks, i.e.,~the attention function does not attend to the indicators.
To test this, we probe the representation after the attention operation at two locations, before and after the residual connection.
Fig.~\ref{fig:linprobe_vit_failure_failure} reveals that the indicator information is available in all layers. Early in training, some indicator information is filtered out in the attention block (blue). 
Indicator information is partially filtered out in deeper layers, but is always recovered by the residual connections (see Fig.~\ref{fig:linprobe_z}).
Thus, features required for task 1 are not filtered out.
\begin{figure}[t]
    \centering
        \includegraphics[width=1\linewidth,trim=0 0 0 0,clip]{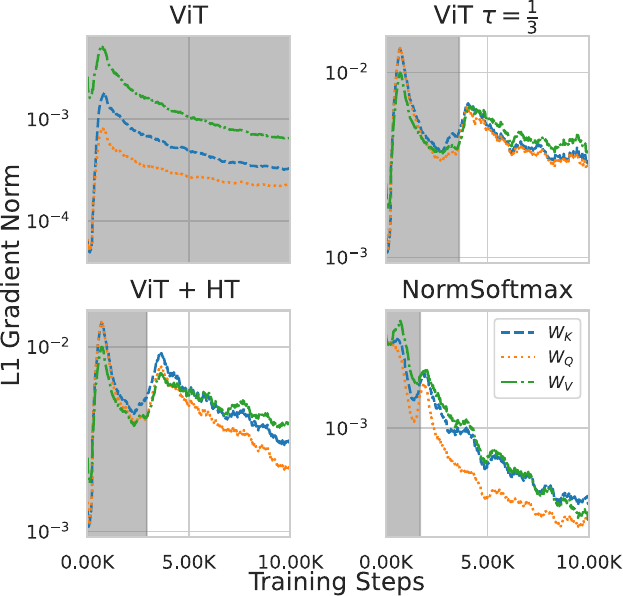}
    \vspace{-1.5\baselineskip}
    \caption{\textbf{L1 gradient norm during training} for $W_K$, $W_Q$ and $W_V$ for the first layer. For ViT, $W_K$ and $W_Q$ receive much smaller gradients than $W_V$. Before \aha (gray regions), the differences between gradient magnitudes are much smaller for smaller temperatures or \normsoftmax. The y-axis is log scaled. All layers shown in Fig.~\ref{fig:grad_mag_full}. }
    \label{fig:grad_mag}
\end{figure}

\begin{figure}
    \begin{subfigure}{0.24\linewidth}
        \footnotesize{ViT}
        \includegraphics[width=1.\textwidth,trim=160 80 230 80,clip]{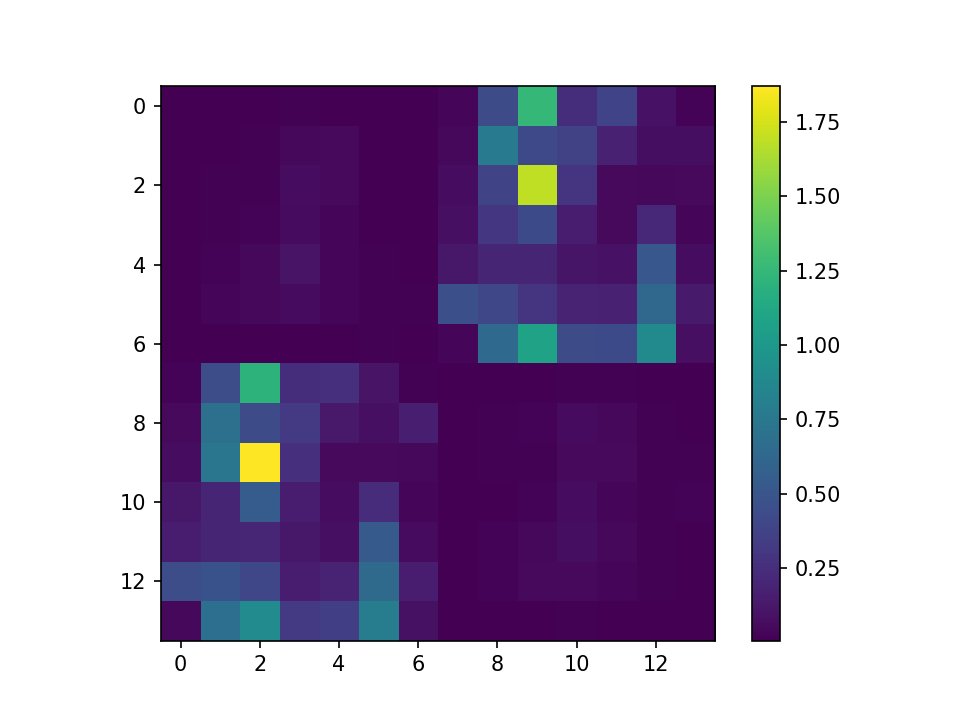}
    \end{subfigure}
    \hfill
    \begin{subfigure}{0.24\linewidth}
        \footnotesize{$\text{ViT }\tau$=1/3}
        \includegraphics[width=1.\textwidth,trim=160 80 230 80,clip]{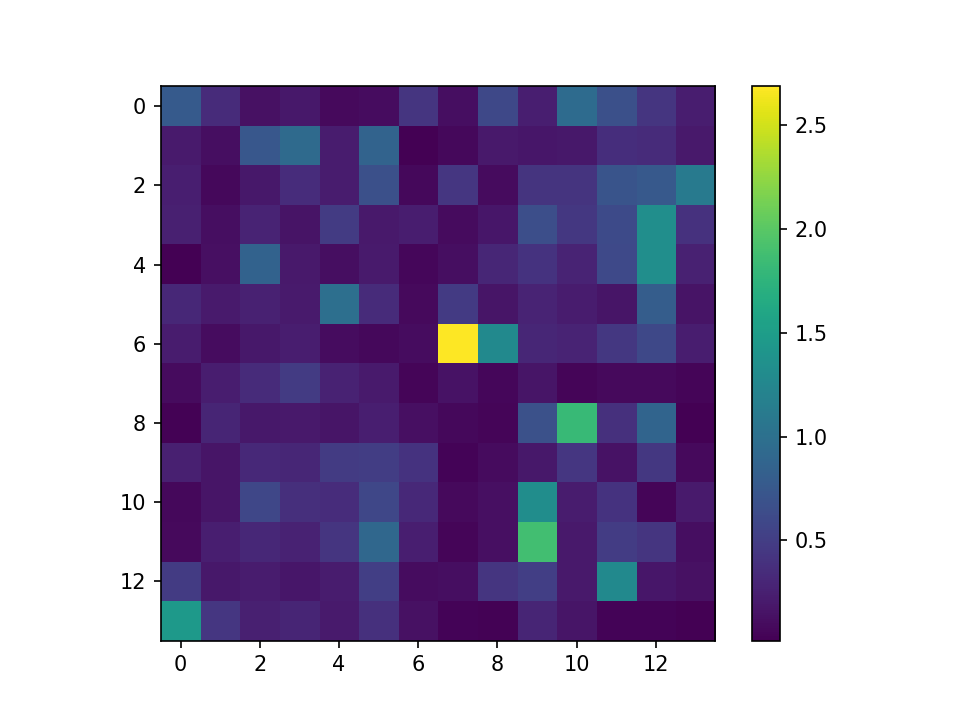}
    \end{subfigure}
    \vspace{-0.5\baselineskip}
    \begin{subfigure}{0.24\linewidth}
        \footnotesize{ViT+HT}
        \includegraphics[width=1.\textwidth,trim=160 80 230 80,clip]{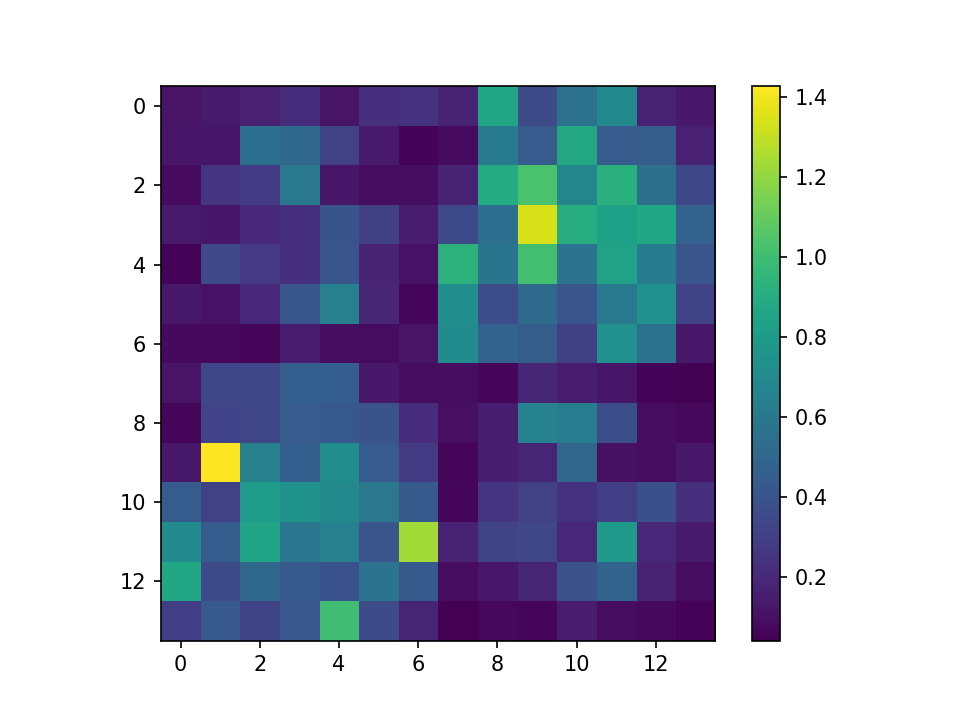}
    \end{subfigure}
    \hfill
    \begin{subfigure}{0.24\linewidth}
        \footnotesize{\normsoftmax}
        \includegraphics[width=1.\textwidth,trim=160 80 230 80,clip]{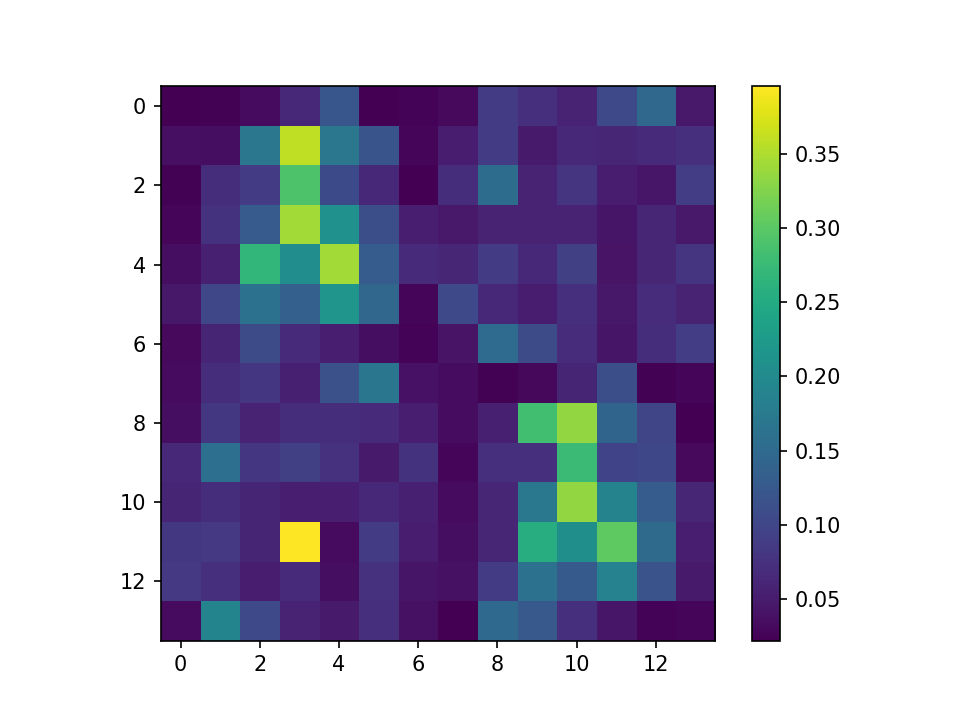}
    \end{subfigure}
    \caption{\textbf{Gradients on image for $W_k$} at Epoch 50. For ViT the gradient for $W_K$ comes mostly from target regions, while for the other approaches indicator regions provide substantial gradient. A detailed explanation of this plot and plots for $Q$ and $V$ can be found in Sec.~\ref{sec:split_grad_indi_target}}
    \label{fig:grad_imgs}
\end{figure}
\begin{table}
  \caption{\textbf{Quantitative results} on the \textbf{main dataset}, as described in Fig.~\ref{fig:teaser_data}. and the \textbf{No position task} (Fig.~\ref{fig:tas}). $\tau$ not optimized for \emph{No position task}. \textbf{ER}: \aharatio, \textbf{Acc.}: Accuracy, \textbf{Avg.~EE.}: average \ahaepoch. }
  \label{tab:final_res_joint}
  \vspace{-1\baselineskip}
  \centering
  \resizebox{1\linewidth}{!}{
  \begin{tabular}{lcccc}
    \toprule
    \multicolumn{5}{c}{Main Dataset}\\
    \midrule
    \multicolumn{2}{c}{} & \multicolumn{1}{c}{} & \multicolumn{2}{c}{Avg.~over EMs}\\
    Model & $\tau$    & ER $\uparrow$    & Acc. $\uparrow$  & Avg.~EE. \\
    \midrule
   ViT &  $\frac{1}{0.025}$ & 3/10 & 89.40 & 174.67 \\
   ViT &  $\frac{1}{0.075}$ & 6/10 & 90.13 &  181.34 \\
   ViT + WD 0.5 &  $\sqrt{d_k}$ & 5/10 & 90.09 & 177.8\\
   ViT &  $\sqrt{d_k}$ & 7/10 & 89.48 & 207.43  \\
   ViT + Warmup 20 &  $\sqrt{d_k}$ & 8/10 & 87.65 & 205.87  \\
   $W_{QKV}$ grad scaling & $\sqrt{d_k}$ & \textbf{10/10} & 87.96 & 119.4 \\ 
    \normsoftmax & $\sqrt{d_k}$ & \textbf{10/10} & 89.56 & 28.2  \\
    \normsoftmax & $\frac{1}{3}$ & \textbf{10/10} & 89.18 & 23.5  \\
    ViT  & $\frac{1}{3}$ & \textbf{10/10} & 89.35 & 66.6  \\
    ViT+HT & $\frac{1}{3} \rightarrow  \sqrt{d_k}$ & \textbf{10/10} & 89.81 & 74.0 \\
    \normsoftmax + HT & $\frac{1}{3} \rightarrow  \sqrt{d_k}$ & \textbf{10/10} & \textbf{89.83} & 17.5 \\
    \midrule
    \multicolumn{5}{c}{No Position Task} \\
    \midrule
   ViT &  $\sqrt{d_k}$ &  0/4 & - & -\\
   ViT + Warmup 20 &  $\sqrt{d_k}$ &  1/4 & 89.55 & 117 \\
   $W_{QKV}$ grad scaling & $\sqrt{d_k}$  & 0/4 & - & - \\ 
    \normsoftmax & $\sqrt{d_k}$ &  \textbf{3/4} & \textbf{88.98} &  228 \\
    \normsoftmax & $\frac{1}{3}$ &  1/4 & 89.77 & 20  \\
    ViT  & $\frac{1}{3}$ &  1/4 & 89.68 & 191 \\
    ViT+HT & $\frac{1}{3} \rightarrow  \sqrt{d_k}$ &  1/4 & 88.36  & 242\\
    \normsoftmax + HT & $\frac{1}{3} \rightarrow  \sqrt{d_k}$ & 1/4  & 90.63   & \textbf{19}  \\    
    \bottomrule
  \end{tabular}
  }
\end{table}

\textbf{Does the transformer fail to combine the information?}
We observe that the target location (solution of task 1) cannot be inferred by the linear probe (Fig.~\ref{fig:linprobe_vit_failure_failure}). Therefore, even though the (indicator) information is available, it is not able to utilize this information to predict the target location of task 2.
Thus, the transformer has all the information but fails to combine it to solve the multi-step task.

\textbf{Differences to a transformer that had a \aha.}
Fig.~\ref{fig:linprobe_vit_failure_success} shows the linear probe results for a transformer that had a \aha. Interestingly, the target location can be inferred by linear probes from all layers and all tested representations with high accuracy. This is in stark contrast to transformers that had no \aha.
The second striking difference is that indicator information is represented in the CLS token. This is even stronger for early layers (Fig.~\ref{fig:linprobe_z}). 
We suspect, that the transformer uses the CLS token to match the classes of the indicators.

\textbf{Why does the transformer fail to combine the information?}
We visualize the attention maps of two fully trained ViTs in Fig.~\ref{fig:attn_maps}. We find that the transformer without \aha does not attend the indicator digits and attends only the targets of task 2, whereas a transformer with \aha attends the indicator digits in early layers.
This suggest that a ViT without \aha does not pay enough attention to the indicators to match them and struggles to learn to attend different regions.

\begin{table}[h!]
  \caption{\textbf{Sensitivity to learning rate}. Lower temperatures and  \normsoftmax drastically increase robustness to imperfect learning rate schedules. \aharatio computed over seeds and schedules.}
  \label{tab:lr_sensitivity}
  \centering
  \vspace{-1\baselineskip}
  \begin{tabular}{lc}
    \toprule
    Model &   \aharatio $\uparrow$  \\
    \midrule
    ViT & 04/36 \\
    VIT + Warmup 20 & 14/36\\
    $W_{QKV}$ grad scaling & 5/36\\
    \normsoftmax & \textbf{36/36} \\
    ViT $\tau=\frac{1}{3}$ & 20/36 \\ 
    ViT+HT $\frac{1}{3} \rightarrow  \sqrt{d_k}$ & 25/36 \\
    \bottomrule
  \end{tabular}
\end{table}

\textbf{Why does the transformer fail to learn to attend to the indicators?}
Based on the discussion in Sec.~\ref{sec:background}, we hypothesize that ill-distributed attention scores lead to small gradients for $W_K$ and $W_Q$, which in turn inhibit learning and in particular learning to shift attention towards the indicators. 
Note that high attention to some pairs with low attention to all others or uniformly distributed attention can result in vanishing gradients \citep{noci2022signal}.
To test this, we visualize the L1-norm of the gradient of the first layer in Fig.~\ref{fig:grad_mag}. 
For vanilla ViT the gradients for $W_K$ and $W_Q$ are 0.5-1 orders of magnitude smaller than those for $W_V$.
Thus, only small gradients are backpropagated through the Softmax to $W_Q$ and $W_K$, and the attention map improves only slowly, which results in the observed slow learning. 
Differences between the gradients are much smaller for ViT $\tau=\frac{1}{3}$, ViT+HT and \normsoftmax, in particular before the \aha. Fig.~\ref{fig:grad_imgs} shows the origin of gradients on the image plane for $W_k$. For ViT, the gradients mostly originate from the target regions, which further explains why many steps are needed to move attention to the indicators, as these gradients  only lead to improved target recognition.

\textbf{Is too small or too large attention-entropy the problem?}
We visualize the distribution of attention maps over training in Fig.~\ref{fig:attn_heatmap}: the vast majority of attention scores is very small. This indicates that a too uniform attention is causing small gradients.
Larger attentions are rare, but not absent, as can be seen in Fig.~\ref{fig:attn_maps_failure}, but indicator regions have small and uniform attention.
Thus, we conclude that \textbf{local uniform attention} causes the transformer's learning problems.

\subsection{Can Enforcing Lower Entropy Attention Maps Resolve the Small Gradients?}

The previous subsection indicates that a local uniform attention is causing the learning problems of the transformer.
To test this hypothesis and show that this causes the observed problems we use targeted interventions.  
Particularly, we modify Softmax's temperature $\tau$ in the attention block.
Large temperatures increase entropy, while small temperatures decrease it.
We apply following interventions: training with lower/higher temperature; HT, where the temperature increases from a low value to default temperature during the first half of training; and \normsoftmax, which adaptively changes the temperature for each sample, head and layer.

\begin{figure}[t]
    \centering
    \includegraphics[width=0.7\linewidth,trim=8 8 8 5,clip]{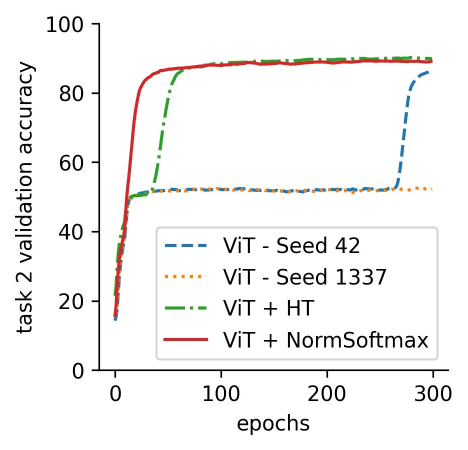}
    \vspace{-0.5\baselineskip}
    \caption{\textbf{Validation accuracy curves on \emph{main dataset}.} Both interventions, ViT+HT and \normsoftmax drastically reduce the saturation period and can even lead to a complete disappearance of the saturation period.}
    \label{fig:final_acc_curve}
\end{figure}

\textbf{Does a lower temperature solve the small gradient issue and thereby mitigate the optimization issues?}
Increasing the temperature from low to default or using \normsoftmax increases high attention scores (c.f.~Fig.~\ref{fig:attn_heatmap}).
Importantly, the transformer learns to also attend to the indicators (Fig.~\ref{fig:attn_maps_success}).
Furthermore, all approaches (lower temperature $\tau$, HT and \normsoftmax) solve the imbalanced gradient issue for $W_V$, $W_Q$ and $W_K$ (Fig.~\ref{fig:grad_mag}) and lead to higher gradients in indicator regions (Fig.~\ref{fig:grad_imgs}).
\ahas happen much earlier or instantly (see Fig.~\ref{fig:final_acc_curve}).
Thus, the interventions indeed mitigate the optimization issues.
A comprehensive comparison between a vanilla ViT and other versions is provided in Tab.~\ref{tab:final_res_joint}.
Decreasing the temperature or using \normsoftmax increases the \aharatio, accuracy and decreases the \ahaepoch (i.e.,~improving the energy landscape).
In contrast, increasing the temperature has a negative effect on the \aharatio, showing that the local uniform attention is the main cause for the learning problem.

\subsection{Is This an Artificial Problem Caused by Other Factors?}
\label{sec:artifical_problem}

\begin{figure}
    \centering
    \includegraphics[width=0.6\linewidth]{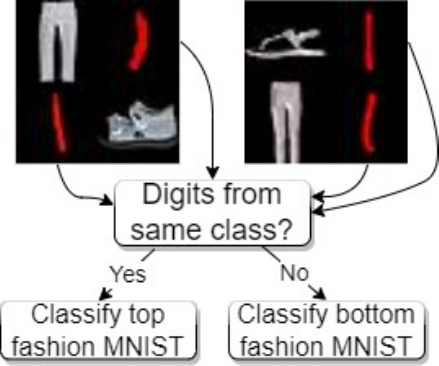}
    \vspace{-0.5\baselineskip}
    \caption{\textbf{\emph{No position task} description.} This task is identical to the \emph{main task}, but removes more information by swapping indicator (digit) and target (fashion) in each row with a probability of 0.5, i.e.,~task 2 is to classify either the top or bottom fashion sample. Two samples are shown to highlight differences from the task described in Fig.~\ref{fig:teaser_data}.}
    \label{fig:tas}
\end{figure}

\textbf{Does the transformer simply ignore specific indicator locations?}
The task and dataset used in the previous subsections showed indicators and targets always at the same location, i.e.,~indicators on top-left and bottom right.
Such a dataset design might result in two undesired effects: 1) The transformer might learn to ignore features based on the associated positional embeddings. 
2) The task might be easier, since positional embeddings can be used as shortcut to find indicators without the need to rely on the actual features.
To disprove both cases we create another dataset, explained in Fig.~\ref{fig:tas}.
We observe that removing the fixed position for indicators and targets makes the task even more difficult (Tab.~\ref{tab:final_res_joint} bottom) and differences between methods are even more apparent. Thus, ViTs without \aha do not simply learn to ignore regions of the image.

\textbf{Is this a mere artifact of a bad choice of hyper-parameters?}
We test various competing explanations that could lead to the observed phenomenon, like training instabilities due to inadequate Warmup, 
bad choice of learning rate schedules and Weight Decay to force circuit formation as observed for grokking \citep{mechanistic_grokking}. 
Weight Decay and Warmup do not improve the results. We find that \ahas are sensitive to the learning rate. \normsoftmax and HT reduce the sensitivity to sub-optimal learning rates drastically (see Tab.~\ref{tab:lr_sensitivity}). See \ref{sec:hyper_parameter_artifact} for a more detailed answer to the posed question.
\begin{figure*}[h!]
    \centering
    \includegraphics[width=0.8\linewidth,trim=8 6 8 3,clip]{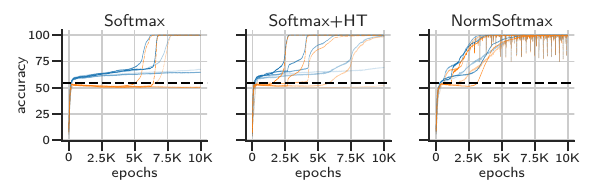}
    \caption{\textbf{\ahas for single-layer transformers on a simple reasoning task.} We show the train (blue) and test (orange) accuracies for attention with Softmax, Softmax+HT, or \normsoftmax, over 5 random seeds (transparencies). Chance probability is \begin{math}6/11\approx 54\%\end{math} (black).}
    \label{fig:non_vision}
\end{figure*}

\begin{figure}[h!]
    \centering
    \begin{subfigure}[c]{1.0\linewidth}
    \centering
    \includegraphics[width=0.7\linewidth,trim=8 0 8 0,clip]{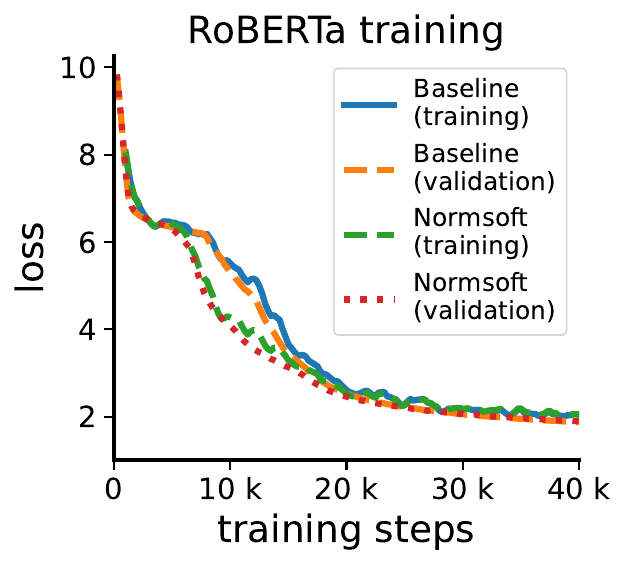}
    \subcaption{Language modeling}
    \end{subfigure}
    \begin{subfigure}[c]{1.0\linewidth}
    \centering
    \includegraphics[width=0.7\linewidth,trim=8 0 8 0,clip]{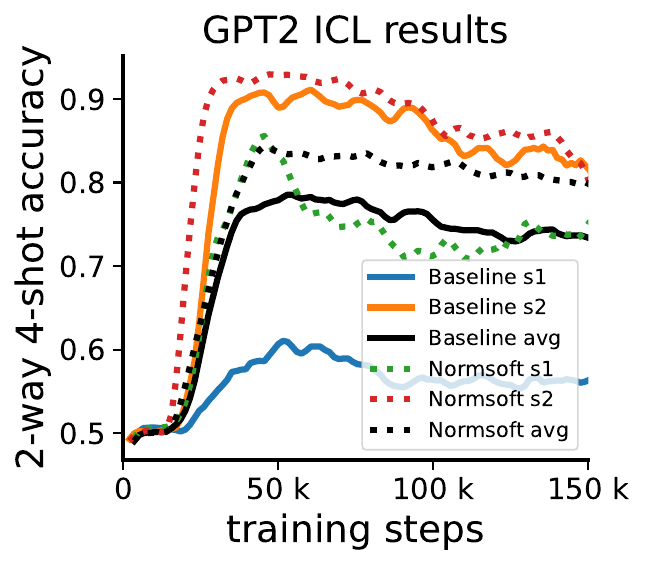}
    \subcaption{In-Context learning}
    \end{subfigure}
    \vspace{-1\baselineskip}
    \caption{\textbf{\ahas on real datasets.} Sharp improvements after plateauing can be observed for RoBERTa pretraining and ICL. Using \normsoftmax leads to an earlier (RoBERTa) and higher (ICL) \aha. Averages for ICL over 4 seeds, 2 exemplary seeds shown per method.}
    \label{fig:real_life_examples_normsoft}
\end{figure}

\textbf{Influence of model scale, dataset size and image resolution on \ahas}.
We found no consistent influence of model scale on \aharatio. See \ref{sec:model_scale} for more details.
Further, we can rule out that this observation is an artifact of image resolution or dataset size by showing the phenomenon on a ImageNet-100 based dataset in \ref{sec:imagenet}.

\textbf{Can the problem be fixed by rescaling of the gradient magnitude for $W_V$, $W_Q$ and $W_K$?}
The observation that lower gradient imbalance leads to higher \aharatio suggests, that simply rescaling of the gradients may solve the problem. 
We find that this does work for the \emph{Main task} but not for the \emph{No position task}, (see Tab.~\ref{tab:final_res_joint}) and is very sensitive to the learning rate (see Tab.~\ref{tab:lr_sensitivity}).
We attribute this to the differences in gradient magnitudes for indicators and targets and discuss it further in Sec.~\ref{sec:grad_mag_scale}.

\textbf{Do gradients vanish completely and can transformers recover?}
Fig.~\ref{fig:teaser_acc} already suggests, that one potential solution to reliably get \ahas is very long training. This observation is supported by Fig.~\ref{fig:grad_mag}, which indicates that gradients become small, but not 0. 
Indeed we observe that training for 3000 epochs results in a \aharatio of 4/4 for all the learning rate schedules.
In practice, this is of little help because the number of sub-tasks is unknown and \ahas are hard to predict.

\textbf{Is this truly a transformer optimization problem?}

To show that \ahas are general to all transformers and not just artifacts of vision data, we also show their occurrence in single-layer 4 head transformers on simplistic algorithmic tasks, referred to as the \textbf{reasoning task} (Sec.~\ref{sec:task_description}). Here, features (numbers) are directly provided as tokens and need not be extracted. 
Fig.~\ref{fig:non_vision} reveals that \ahas appear even in this minimal setting.
Both HT and \normsoftmax reduce the training steps required for \ahas to occur and increase the \aharatio from 3/5 to 4/5 or 5/5, respectively. 

\textbf{Does the \normsoftmax intervention translate to real data?}
\citet{normsoftmax} reported improved performance and faster convergence on ImageNet and machine translation tasks using \normsoftmax.
Both tasks likely contain some innate multi-step tasks, e.g.,~identifying a common discriminative feature and then discriminating between the difficult classes for ImageNet. These improvements may be due to easier multi-step learning with \normsoftmax. 

To further validate the results of our analysis in a real setting we train RoBERTa \citep{roberta} with \normsoftmax for language modeling on Wikipedia, where we suspect task 1 to be learning of general word probabilities and task 2 to be modulating these probabilities based on context, or modulating them based on syntactic relations \citep{chen2024sudden}.
Additionally, we train GPT-2 on the Omniglot \citep{radford2019language} dataset and test its ICL abilities following \citet{chan2022data}. Experimental details for RoBERTa and ICL are provided in ~\ref{sec:roberta_details} and ~\ref{sec:gpt2_details}, respectively.
Fig.~\ref{fig:real_life_examples_normsoft} shows that \normsoftmax indeed leads to earlier \ahas for RoBERTa. 
Furthermore, \normsoftmax also improves ICL, i.e., GPT-2 with \normsoftmax trained on the Omniglot ICL task results in higher ICL accuracy and seems to prevent failure cases like no or a small \aha (Baseline s1 in Fig.~\ref{fig:real_life_examples_normsoft}).
Thus, our analysis and results are indeed transferable to real datasets and also to transformers with causal attention,i.e., GPT-2. 

\section{Limitations and Conclusion}

\textbf{Limitations.}
The ability to decompose tasks into sub-problems and learn to solve those sub-tasks is a common problem, but it is difficult to study on real datasets, due to many confounding factors. 
As a result many works follow  a trial-and-error approach. In contrast, we try to gain deeper understanding by studying this problem in a controlled, synthetic setting. This comes with the assumption that our analysis transfers to real data. We find evidence for that, as the intervention (\normsoftmax) leads to improvements on real data and can prevent getting stuck (Fig.~\ref{fig:real_life_examples_normsoft}).

\textbf{Conclusion.}
In this work, we identified that transformers have difficulties to decompose a task into sub-problems and learn to solve the intermediate sub-tasks. We observed that transformers can learn these tasks suddenly and unexpectedly but usually take a long time to do so. We called these sudden leaps \ahas. We pined the problem down to the Softmax in the attention that leads to small gradients. 
We proposed simple interventions that specifically target the Softmax and show that they improve the transformers' capabilities to learn sub-tasks and to learn them faster. We identify \normsoftmax as most robust and convenient method, leading to consistently better results. Last, we showed that our observations transfer to real datasets.

\section*{Impact Statement}
We do not see any particular societal impact other than the overall impact of advancing the field of Machine Learning.

\section*{Acknowledgements}
We thank Philipp Schröppel and the entire Computer Vision Group Freiburg for inspiring discussions.
Sudhanshu Mittal for feedback on the manuscript and Matteo Farina for reporting observations of
potential Eureka-moments on vision and language tasks.

This research was funded by the Bosch Center for Artificial Intelligence, the Bundesministerium für Umwelt, Naturschutz, nukleare Sicherheit und Verbraucherschutz (BMUV, German Federal Ministry for the Environment, Nature Conservation, Nuclear Safety and Consumer Protection) based on a resolution of the German Bundestag (67KI2029A), the Deutsche Forschungsgemeinschaft (DFG, German Research Foundation) – SFB 1597 – 499552394 and 417962828.

\bibliography{refs.bib}
\bibliographystyle{icml2024}

\clearpage
\appendix
\section{Appendix}
\begin{figure*}[h!]
    \centering
    \includegraphics[width=1.0\textwidth]{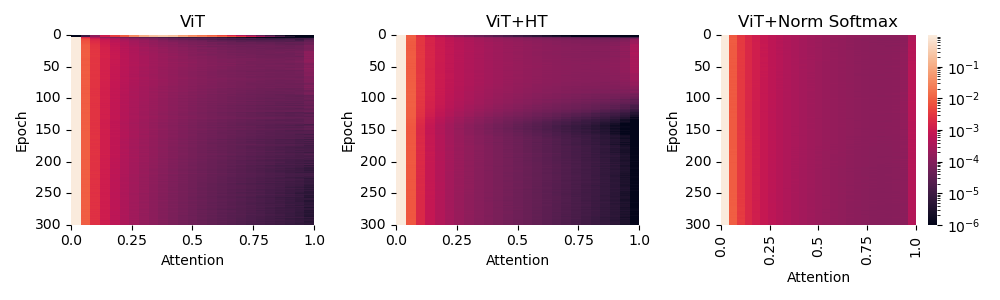}
    \caption{\textbf{Attention distribution as a heatmap. } Attention scores are sampled during evaluation after each epoch and binned to 25 bins. The color map is log scaled. For all 3 models, the vast majority of values falls into the first bin. ViT shows very few higher attention scores. ViT+HT and ViT+\normsoftmax lead to a significantly larger number of medium and high attention values. For ViT+HT this is limited to the first 100 epochs.}
    \label{fig:attn_heatmap}
\end{figure*}
\begin{figure*}[h!]
    \vspace{4\baselineskip}
    \centering
    \includegraphics[width=1.0\textwidth]{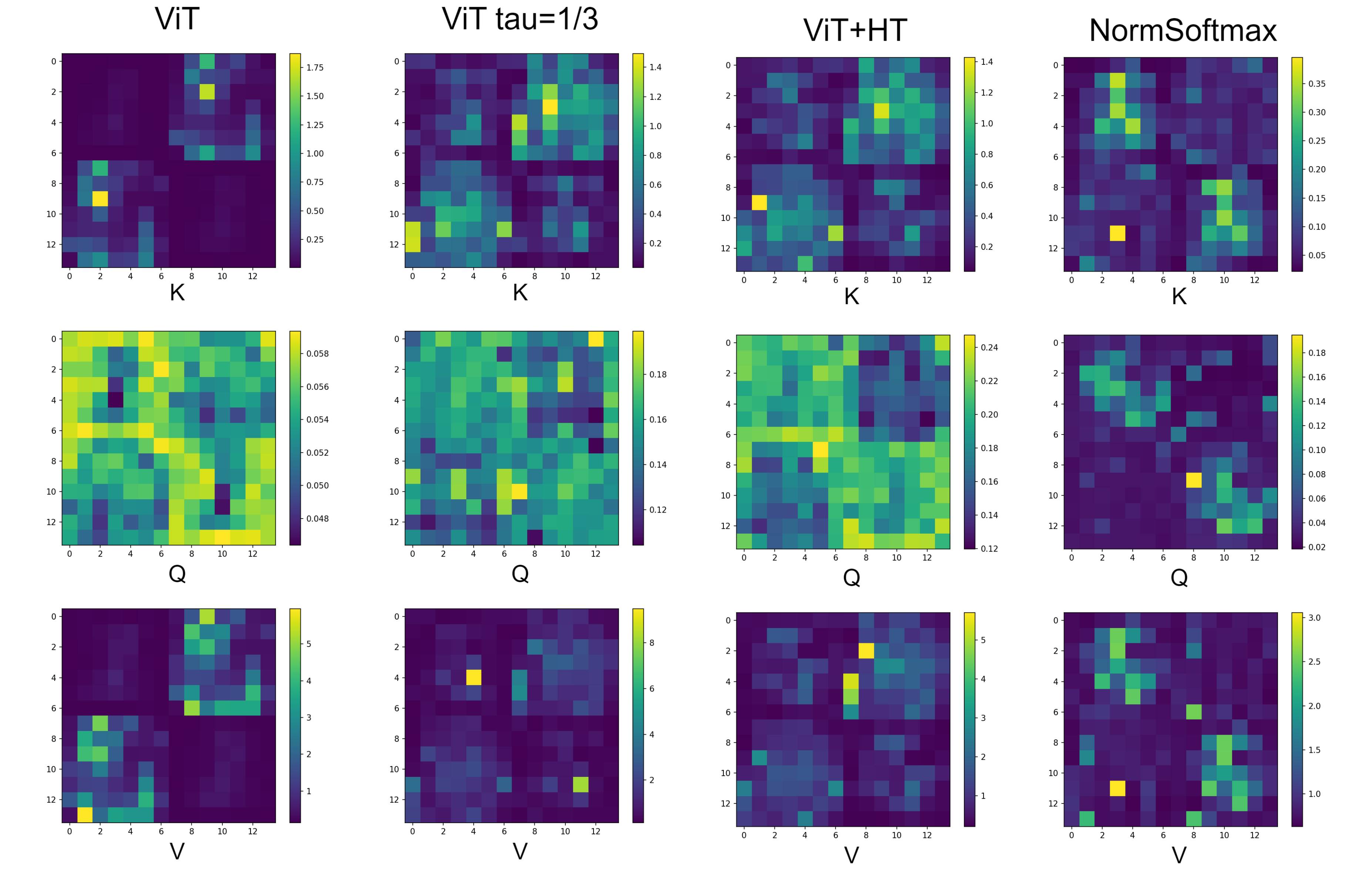}
    \caption{\textbf{Gradient norm for different image regions visualized for $Q$, $K$, $V$.} Larger gradients are visible for target regions, i.e,~top-right and bottom-left. Indicator receive less gradients for $K$ and $V$. ViT $\tau=\frac{1}{3}$, ViT+HT and \normsoftmax mitigate this problem well. Indicator regions for $Q$ receive more gradient relative to the target regions, but overall the gradient for $Q$ is very small (see color bar). Plots created at epoch 50.}
    \label{fig:grad_map}
\end{figure*}

Here, we provide additional information that supports understanding and helps interpreting the main paper. 
We provide supplemental experimental results and more detailed analysis. We show the gradient norm for indicators and targets individually, which reveals that most of the already small gradients for $W_Q$ and $W_K$ is attributed to target features for models that do not learn the task and very little to indicator features.
We provide a more complete version of Fig.~\ref{fig:grad_mag}.
Provide more insights into why gradient magnitude scaling does not work and provide more details on our analysis of competing hypotheses.
We provide an ablation on model scale and the main results from the main paper with standard deviation and training speed up.
We explain additional datasets and report results on them. 
We repeat the linear probe analysis with more sensitive information probes.
We provide the linear probe plots using also $Q$, $K$ and $V$ representations and the full version of Fig.~\ref{fig:grad_mag}.
Last, we provide details on training and experimental setups and an explanation for vanishing gradients in case of centralized attention maps.

\subsection{Attention Distribution Over Time}

We show the change of the attention distribution during the course of training in \cref{fig:grad_map}
\begin{figure*}[h]
    \centering
    \includegraphics[width=1.0\textwidth]{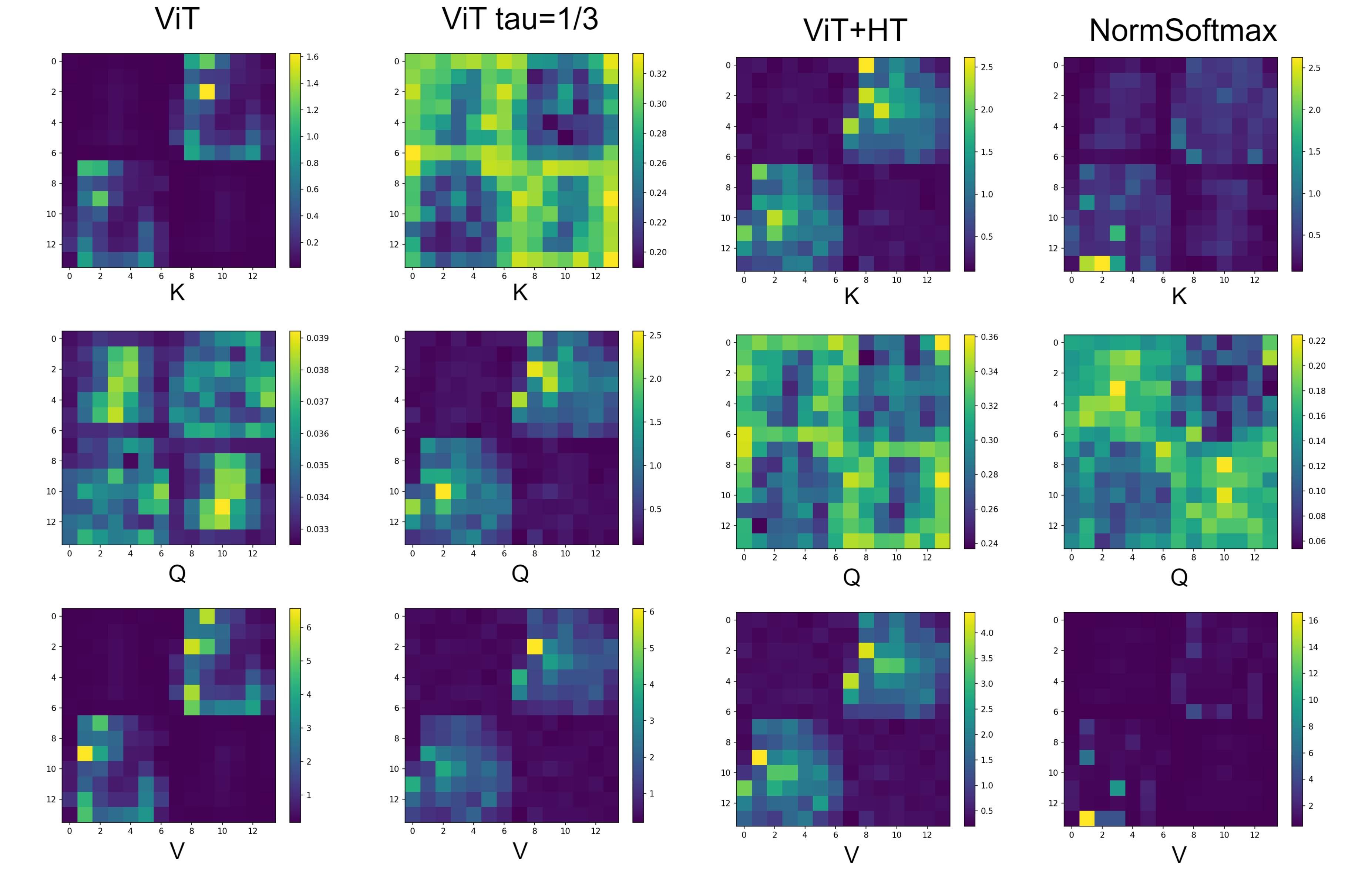}
    \caption{\textbf{Gradient norm for different image regions visualized for $Q$, $K$, $V$ }at epoch 13 (\aha of \normsoftmax). }
    \label{fig:grad_map_ep13}
\end{figure*}

\subsection{Gradient Norm for Indicators and Targets}
\label{sec:split_grad_indi_target}
Our particular dataset design allows us to look at the gradients for indicators and targets separately. In particular, we make use of the fact, that indicators and targets are always at the exact same spacial location. More precisely, we use the partial derivatives as proxy for the gradients. We compute $\frac{\partial Z}{\partial Q}$, $\frac{\partial Z}{\partial K}$ and $\frac{\partial Z}{\partial V}$, where $Z$ is the output of the attention function. 
To analyze the gradient norm for targets and indicators independently we compute $\frac{\partial Z}{\partial Q}$, $\frac{\partial Z}{\partial K}$ and $\frac{\partial Z}{\partial V}$, where $Z$ is the output of the attention function.
Since we compute the derivative wrt.~the tokens, the spacial dimension remains. 
By averaging over the batch dimension and heads we can plot the partial derivative for each token. While it's not exactly the same, we will use the term gradient to refer to these partial derivatives in the following.

Since each token corresponds to a region in the image, we can visualize these results as an image. The results are shown in Figs.~\ref{fig:grad_imgs}, \ref{fig:grad_map}, \ref{fig:grad_map_ep13}. 
It can clearly be seen, that target regions (top-right and bottom-left) receive more gradients than indicator regions for $K$ and $V$. ViT, ViT+HT and \normsoftmax mitigate this problem, leading to significant gradient for indicator tokens.
Indicator regions for $Q$ receive comparatively larger gradients, however, the gradients for $Q$ are much smaller.

\begin{figure*}[h!]
    \centering
    \includegraphics[width=0.85\textwidth]{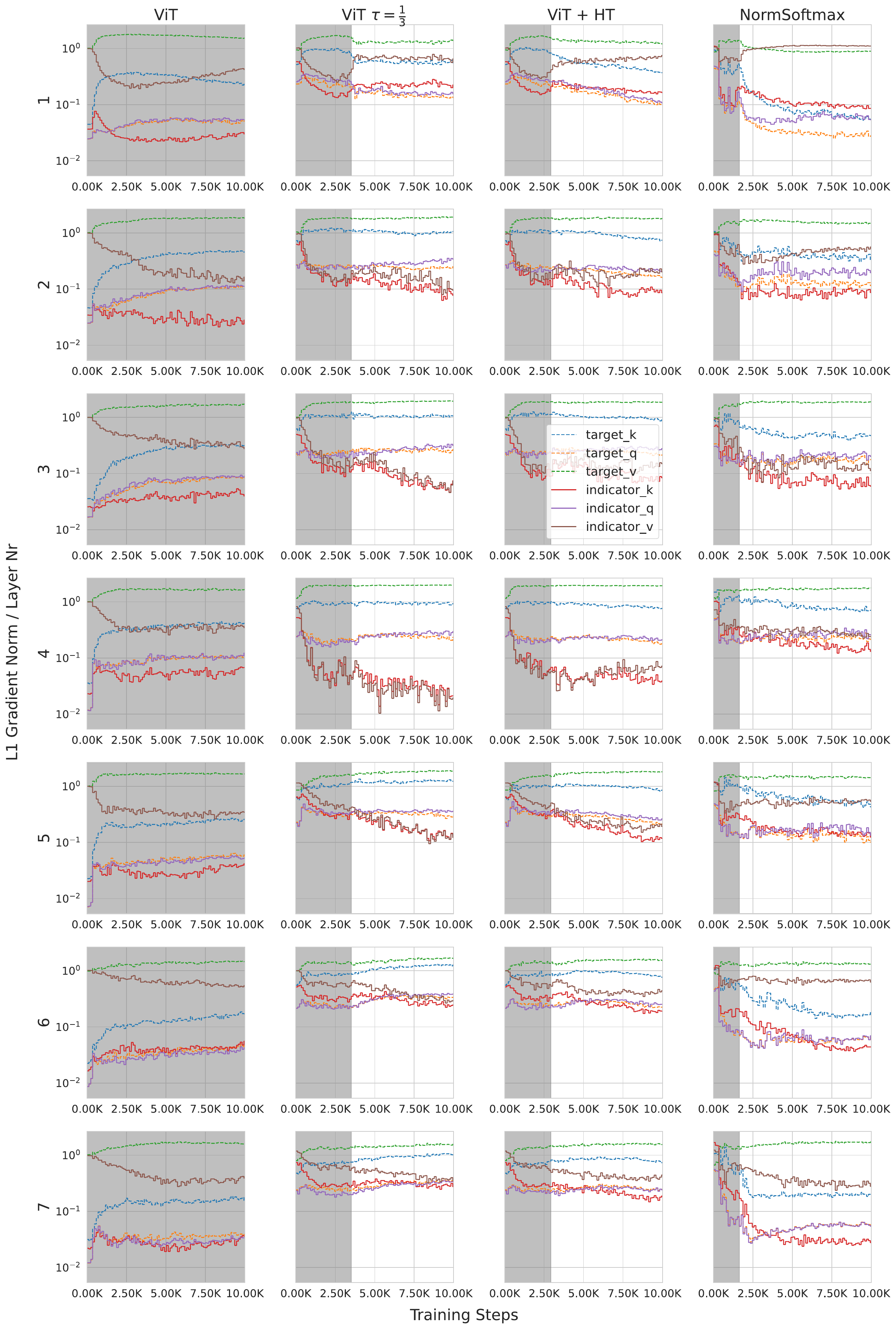}
    \caption{\textbf{L1 gradient norm separately for indicator and target tokens.} Indicator regions/features receive much less gradient than target regions. Gray region indicates steps before \aha.}
    \label{fig:grad_mag_indi_target_split}
\end{figure*}

Besides that, we compute the mean partial derivative for indicator and target regions of the image, i.e.,~we average the partial derivatives for tokens corresponding to target regions or indicator regions. This allows us to plot the gradient norm for $Q$, $K$ and $V$ for only target and indicator tokens over the training. We show the results in Fig.~\ref{fig:grad_mag_indi_target_split}.
We make the following observations:
\begin{enumerate}
    \item In general, the gradients are not evenly divided between target and indicator $K$, with usually smaller gradient for the indicator regions. Therefore Fig.~\ref{fig:grad_mag} even underestimates the difference for the indicator regions, i.e.,~the regions relevant for a \aha. 
    
    \item This difference can explain why more time is needed to reach a \aha for ViT in comparison to the other methods.

    \item This difference between indicator $k$ and target $k$ can also explain, why the $W_{QKV}$ grad scaling does not solve the problem.

    \item For ViT the difference between gradients of ``$V$'' and $K$'' and ``$V$'' and ''$Q$'' is, for most layers, generally much larger compared to the other approaches. This is particularly true before the \aha, where larger gradients for indicator $K$ and indicator $Q$ are crucial to get an early \aha.
    Most prominent is the difference between target $V$ and indicator $K$, showing a large mismatch. This explains, why the attention maps change so slowly and why simply increasing the learning rate does not solve the problem.
\end{enumerate}

\subsection{Gradient Norm}
\begin{figure*}[h!]
    \includegraphics[width=1\textwidth]{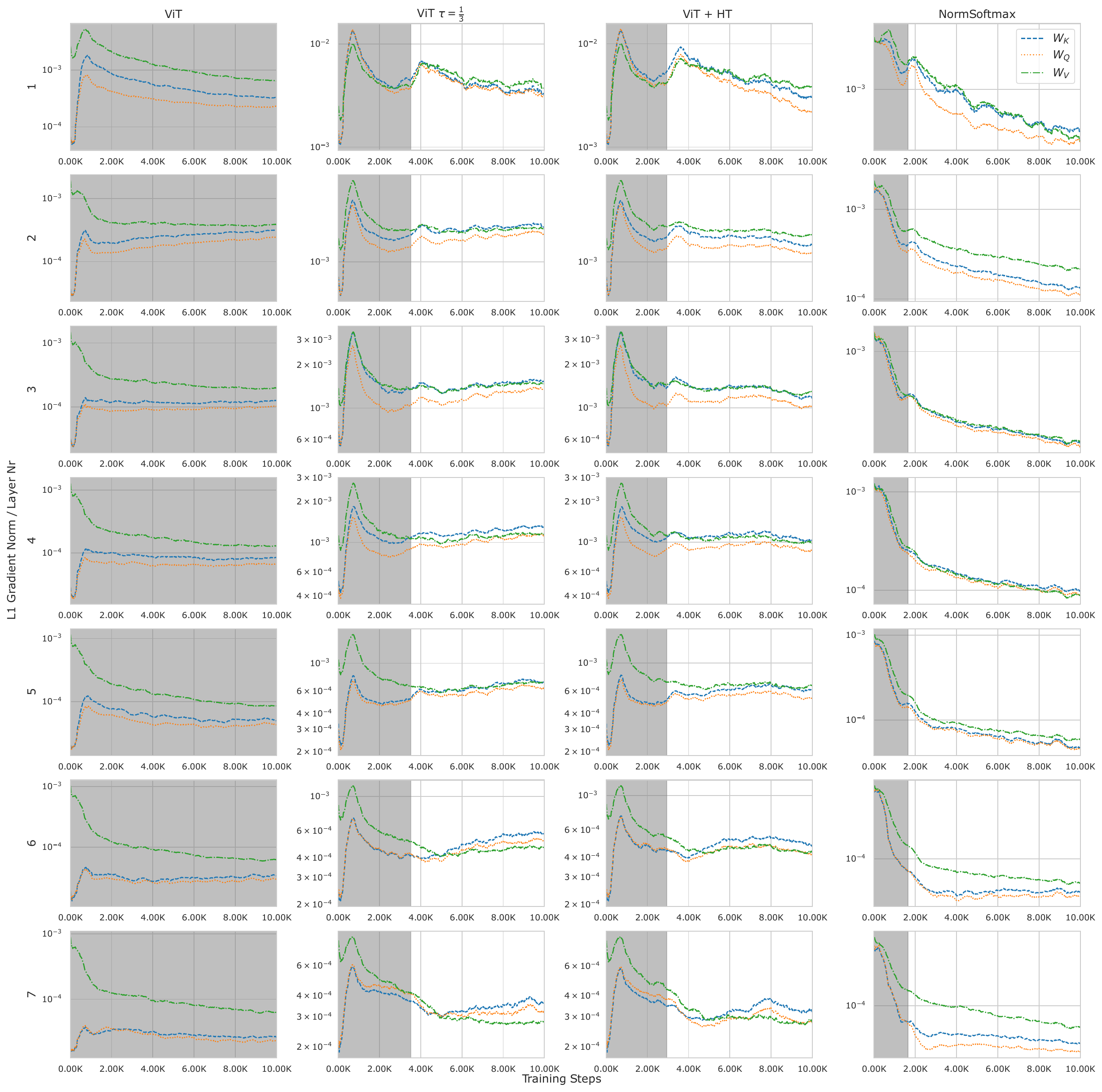}
    \caption{\textbf{L1 gradient norm plot for all layers, all models and entire training.} This is the complete version of Fig.~\ref{fig:grad_mag}. For ViT it can be seen that for all layers the gradient for $W_V$ is significantly larger than for $W_Q$ and $W_K$. For ViT+HT, ViT with $\tau=\frac{1}{3}$ and \normsoftmax the gradient norm is very similar for the weight matrices (note that the y-axis is not shared). \normsoftmax achieves this also for deeper layers. Often after \aha $W_V$ starts to get larger attention than $W_K$ and $W_Q$. We conjecture that this is because the attention is already optimized, while task 2 can still improve by modifying the feature representation. Gray region indicates steps before \aha.}
    \label{fig:grad_mag_full}
\end{figure*}

Fig.~\ref{fig:grad_mag_full} shows the L1 gradient norm for all layers and all methods. It can be seen, that throughout all layers and the entire training ViT has larger gradients for $W_V$ in comparison to $W_K$ and $W_Q$.
For the other methods differences are much smaller and less consistent. 

\subsection{Why Gradient Magnitude Scaling Does Not Work.}
\label{sec:grad_mag_scale}
Following the observation of Fig.~\ref{fig:grad_mag}, it stands to reason to simply scale the gradients for $W_V$, $W_Q$ and $W_K$ to the same value. To this end, we compute the gradient norm for $W_V$, $W_Q$ and $W_K$ for each layer and the overall mean. We scale the gradients for $W_V$, $W_Q$ and $W_K$, such that their norm is equal to the mean norm. This removes the imbalanced gradient issue and results in identical effective learning rate.

While this approach might work, it solves only part of the problem. Different features might receive differently large gradients. For instance the indicator features (here digits) receive little gradient, while target (here fashion) receive large gradients, as can be seen in Fig.~\ref{fig:grad_map}. Simply scaling up the gradients would not solve the imbalance between indicator and target gradients.

We can see in Tab.~\ref{tab:final_res_joint}, that $W_{QKV}$ grad scaling helps  on the main dataset, but is very sensitive to the learning rate (Tab.~\ref{tab:lr_sensitivity}). However, it completely fails on the harder task. The learning rate sensitivity and the failure on the harder task are most likely due to the gradient imbalanced discussed above.

\subsection{Are the Learning Problems a mere Artifact of a Bad Choice of Hyper-parameters? }
\label{sec:hyper_parameter_artifact}
We always use the default of 5 Warmup epochs to avoid training instabilities during early stages of training.
We found that 20 Warmup epochs were most effective in mitigating the problem. However, sensitivity to the learning rate schedule (Tab.~\ref{tab:lr_sensitivity}) is high. 
The average \ahaepoch is very late  (Tab.~\ref{tab:final_res_joint} left) and we found more Warmup epochs lead to worse results on harder tasks (Tab.~\ref{tab:final_res_joint} right).
The \aharatio is sensitive to the learning rate schedule. We test 9 learning rate schedules for each method, (Tab.~\ref{tab:lr_schedule}). Lower temperatures are less sensitive to the learning rate schedule (see Tab.~\ref{tab:lr_sensitivity}).

Weight Decay (WD) can facilitate grokking \citep{grokking} by forcing the network to learn general mechanisms \citep{grokking,mechanistic_grokking}.
In our setting, we only found mild improvements for higher WD. However, more random seeds revealed, that higher WD rather reduces the \aharatio (Tab.~\ref{tab:final_res_joint}) and does not help in solving transformer's learning issue.

\subsection{Influence of Model Scale on \aharatio.}
\label{sec:model_scale}
The low \aharatio could also be due to a too large or too small architecture. Also the number of heads might play an important role, since different features can be attended in different heads. More heads might increase the likelihood of one head specializing in indicators.
Also the embedding dimension per head might just be too small or large for the task at hand. Maybe, even the hidden dimension of the MLP at the end of the attention block is the bottleneck. 
Many parameters of the transformer itself could explain why it fails to solve our tasks. We test these hypotheses in Tab.~\ref{tab:vit_hp}. 
While most changes lead to a lower \aharatio, reducing the depth and increasing the number of heads leads to mild improvements. 
Combining both leads to a \aharatio of 4/4, but, as can be seen in Tab.~\ref{tab:final_res_no_pos_with_std}, this architecture does not generalize to other datasets.

\begin{table*}
  \caption{\textbf{Influence of model scale on \aharatio}. \aharatio is only partially influenced by the architecture. More shallow models and more heads both improve the results. The combination leads to 4/4 \aharatio, but as can be seen in Tab.~\ref{tab:final_res_no_pos_with_std} this architecture fails at other tasks, while our solutions lead to improvements even on the ``no position task''. }
  \label{tab:vit_hp}
  \centering
  \begin{tabular}{cccc|cccc}
    \toprule
    \multicolumn{5}{c}{} & \multicolumn{2}{c}{Avg. over subset with \aha} \\
    \cmidrule(lr){6-7}
    Heads & Emb. Dim. & Depth & MLP & \aharatio $\uparrow$    & Accuracy $\uparrow$  & Avg.~\ahaepoch  \\
    \midrule
      4 & 64 & 7 & 2 & 2/4  & 88.61 $\pm$ 1.64 & 232.50 $\pm$ 35.50\\
      4 & 64 & 7 & 4 &1/4 & 90.16 $\pm$ 0.00 & 162.00  $\pm$ 0.00 \\
      4 & 64 & 4 & 2 & 3/4 & 88.77 $\pm$ 1.20 & 216.33 $\pm$ 46.64\\
      4 & 64 & 10 & 2 & 2/4 & 89.93 $\pm$ 0.26 & 221.00 $\pm$ 18.00\\
      4 & 48 & 7 & 2 & 1/4 & 90.18 $\pm$ 0.00  & 137.00 $\pm$ 0.00 \\
      4 & 96 & 7 & 2 & 2/4 & 89.76 $\pm$ 0.03  & 140.50 $\pm$ 70.5\\
      2 & 64 & 7 & 2 & 0/4 & - & - \\
      6 & 64 & 7 & 2 &3/4 & 89.67 $\pm$0.25 & 139.67 $\pm$ 55.16\\
      6 & 64 & 4 & 2 & \textbf{4/4} & \textbf{89.75 $\pm$ 0.24}  & 152.25 $\pm$ 44.49\\
    \bottomrule
  \end{tabular}
\end{table*}

\subsection{Transformers Learn the Prior $p(z)$}
\begin{figure*}
    \centering
    \includegraphics{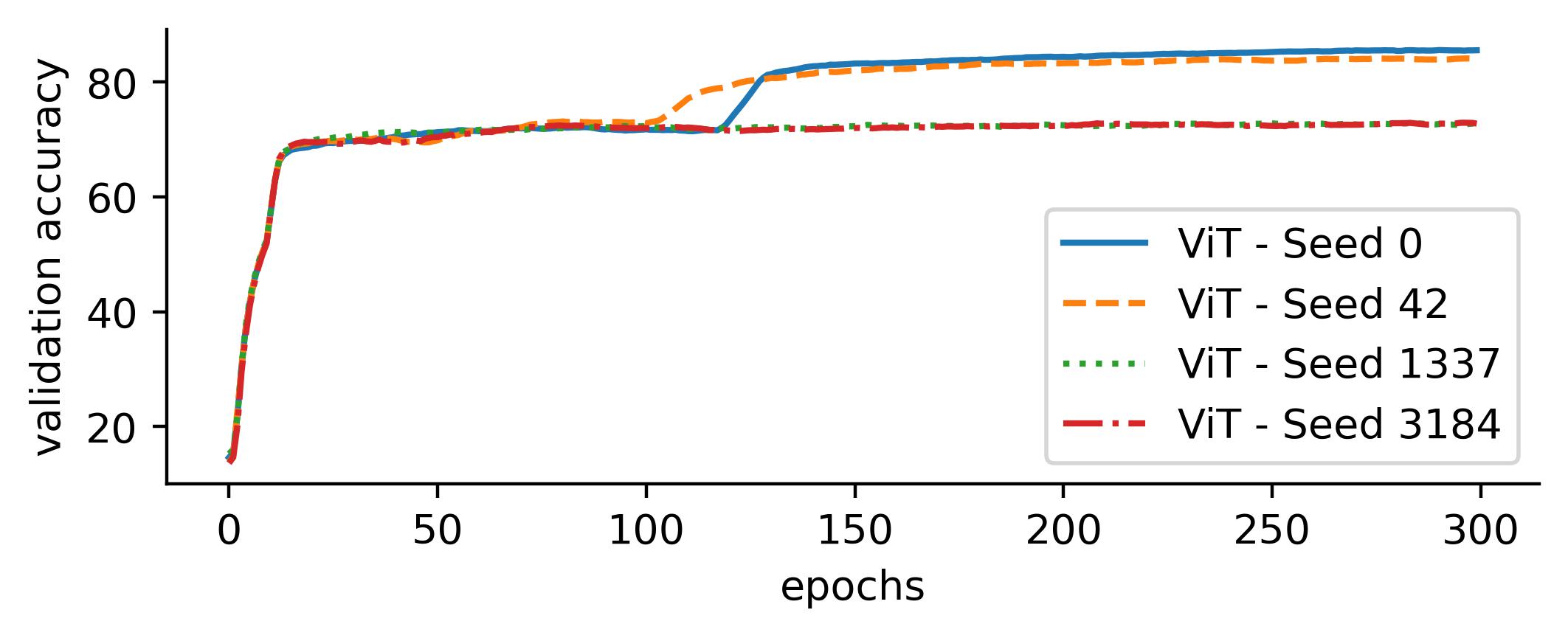}
    \caption{\textbf{Validation accuracy curves for ``main dataset'' with changed target probabilities.} ViT learns that one target is more likely than the other and learn to always pick this target, as can be seen by the higher plateau accuracy.}
    \label{fig:val_acc_065}
\end{figure*}

Given a task like $p(y|x,z) \cdot p(z|x)$, i.e.,~the probability of class $y$ given evidence $x$ and the latent variable $z$, we argue, that transformers first learn a prior $p(z)$, ignoring the evidence. Sometimes they fail to unlearn this and pay attention to the evidence.
In all previous experiments, the probability of target 1 or target 2 being the target to classify was 0.5.
In a setting without 0.5 probability, the transformer should pick the target which is more frequently correct, in case it actually learns the prior $p(z)$ . 
We test this by changing the probability of the top-right target to be the target location to 0.65.
As can be seen in Fig.~\ref{fig:val_acc_065}, the transformer initially learns the shortcut of always picking the more likely target. 

\subsection{Main Results with Standard Deviation, ResNet and Convergence Speed Improvements}
\begin{table*}[h!]
  \caption{\textbf{Main dataset -- Comparison of proposed solutions and baselines.}  This is a complete version of Tab.~\ref{tab:final_res_joint} including standard deviation, and speed improvements. For the \textbf{main dataset}, as described in Fig.~\ref{fig:teaser_data}. \textbf{ER}: \aharatio, \textbf{Acc.}: Accuracy, \textbf{Avg.~EE.}: average \ahaepoch. \textbf{\% of steps} indicates the \% of steps needed to reach 95\% of ViTs accuracy. \textbf{95\%-ratio} indicates the ratio of models that actually reached 95\% of ViTs accuracy. }
  \label{tab:final_res_joint_with_std}
  \centering
  \begin{tabular}{lc|rrrrr}
    \toprule
    & & \multicolumn{5}{c}{Main Dataset}\\
    \midrule
    \multicolumn{2}{c}{} & \multicolumn{1}{|c}{} & \multicolumn{2}{c}{Avg.~over EMs} & \multicolumn{2}{c}{Avg.~over ViT 95\% Acc.}\\
    \cmidrule(lr){4-5} \cmidrule(lr){6-7} 
    Model & $\tau$    & ER $\uparrow$    & Acc. $\uparrow$  & Avg.~EE. & \% of steps & 95\%-ratio \\
    \midrule
    ResNet & & 10/10 & 99.40 $\pm$ 0.10 &  3.00 $\pm$ 00.00 & - & -\\
   ViT &  $\frac{1}{0.025} $ & 3/10 & 89.40 $\pm$  0.08 & 174.67 $\pm$ 37.82 & 84.26 & 3/10 \\
   ViT &  $\frac{1}{0.075}$ & 6/10 & 90.13 $\pm$ 0.40 &  181.34 $\pm$ 24.94 & 86.79 &2/10  \\
   ViT + WD 0.5 &  $\sqrt{d_k}$ & 5/10 & 90.09 $\pm$ 0.27 & 177.80 $\pm$ 52.30 & 84.70 & 5/10 \\
   ViT &  $\sqrt{d_k}$ & 7/10 & 89.48 $\pm$ 1.10 & 207.43 $\pm$ 46.65 & 100.00 & 4/10  \\
   ViT + Warmup 20 &  $\sqrt{d_k}$ & 8/10 & 87.65 $\pm$  6.48  & 205.87 $\pm$ 57.05 & 91.87 & 5/10  \\
    $W_{QKV}$ grad scaling & $\sqrt{d_k}$ & \textbf{10/10} & 87.96 $\pm$ 2.45 & 119.4 $\pm$ 62.70 & 73.79 & 5/10 \\ 
    \normsoftmax & $\sqrt{d_k}$ & \textbf{10/10} & 89.56 $\pm$ 0.65 & 28.20 $\pm$ 34.85 & 19.87 & 10/10 \\
    \normsoftmax & $\frac{1}{3}$ & \textbf{10/10} & 89.18 $\pm$ 0.36 & 23.50 $\pm$ 08.15 & 19.25 & 10/10 \\
    ViT  & $\frac{1}{3}$ & \textbf{10/10} & 89.35 $\pm$ 0.28 & 66.60 $\pm$ 58.55 & 36.71 & 10/10\\
    ViT+HT & $\frac{1}{3} \rightarrow  \sqrt{d_k}$ & \textbf{10/10} & \textbf{89.81} $\pm$ 0.29 & 74.00 $\pm$ 61.29 & 39.88 & 10/10\\
    \normsoftmax + HT & $\frac{1}{3} \rightarrow  \sqrt{d_k}$ & \textbf{10/10} & 89.83 $\pm$ 0.41 & 17.50 $\pm$ 04.84 & 16.41 & 10/10 \\
    \bottomrule
  \end{tabular}
\end{table*}
\begin{table*}[h!]
  \caption{\textbf{No Position Task -- Comparison of proposed solutions and baselines.} For the \textbf{No position task}, as described in Fig.~\ref{fig:tas}. $\tau$ not optimized for this task. \textbf{ER}: \aharatio, \textbf{Acc.}: Accuracy, \textbf{Avg.~EE.}: average \ahaepoch. }
  \label{tab:final_res_no_pos_with_std}
  \centering
  \begin{tabular}{lc|rrr}
    \toprule
    & & \multicolumn{3}{c}{No Position Task}\\
    \midrule
    \multicolumn{2}{c}{} & \multicolumn{1}{|c}{} & \multicolumn{2}{c}{Avg.~over EMs}    \\
    \cmidrule(lr){4-5} 
    Model & $\tau$    & ER $\uparrow$    & Acc. $\uparrow$  & Avg.~EE. \\
    \midrule
    ResNet & & 4/4 & 91.27 $\pm$ 0.30 & 4.25 $\pm$ 00.43\\
   ViT &  $\sqrt{d_k}$  & 0/4 & - & -\\
   ViT + Warmup 20 &  $\sqrt{d_k}$ & 1/4 & 89.55 $\pm$ 0.00 &  117 $\pm$ 00.00 \\
    $W_{QKV}$ grad scaling &  $\sqrt{d_k}$ & 0/4 & - & - \\
    \normsoftmax & $\sqrt{d_k}$  & \textbf{3/4} & \textbf{88.98} $\pm$ \textbf{0.55} & \textbf{228.67} $\pm$ 08.22 \\
    \normsoftmax & $\frac{1}{3}$ & 1/4& 89.77 $\pm$ 0.00 & 20.00 $\pm$ 00.00\\
    ViT  & $\frac{1}{3}$ & 1/4 & 89.68 $\pm$ 0.00 & 191.00 $\pm$ 00.00 \\
    ViT+HT & $\frac{1}{3} \rightarrow  \sqrt{d_k}$ & 1/4 & 88.36 $\pm$ 0.00  & 242.00 $\pm$ 00.00\\
    \normsoftmax + HT & $\frac{1}{3} \rightarrow  \sqrt{d_k}$ & 1/4 & 90.63 $\pm$ 0.00 & 19.00 $\pm$ 00.00 \\
    ViT 6 heads, depth 4 &  & 0/0 & - & - \\
    \bottomrule
  \end{tabular}
\end{table*}

Due to space and readability constraints we report in the main paper only the mean over all seeds. In Tabs.~\ref{tab:final_res_joint_with_std} \& \ref{tab:final_res_no_pos_with_std} we show the same tables including the standard deviation. 

Additionally, Tabs.~\ref{tab:final_res_joint_with_std} \& \ref{tab:final_res_no_pos_with_std} also provide a comparison to a ResNet9. 

Lastly, for Tab.~\ref{tab:final_res_joint_with_std} we report the improved convergence speed as a percentage of the number of training steps to reach 95\% of ViT accuracy (averaged only over seeds with \ahas), denoted as ``\% of steps''. This value is computed only over the fraction of seeds, that actually lead to a higher accuracy than 95\% of ViTs accuracy. 
In the last column, we also report this fraction. Note, that the \aharatio is the maximum possible value for the ``95\%-ratio'', i.e.,~for ``ViT +Warmup 20'' 8/10 seeds have a \aha. Out of these 8 only 5 reach an accuracy higher than 95\% of the ViT accuracy.

\subsection{Description of and Results on More Datasets}
\label{sec:more_datasets}

\begin{figure*}
    \centering
    \begin{subfigure}[b]{0.3\linewidth}
    \includegraphics[width=1.0\textwidth]{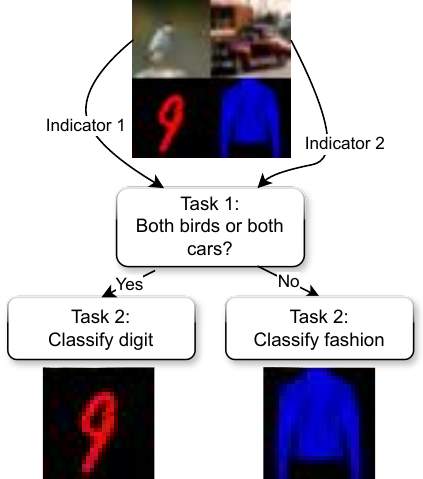}
    \caption{\textbf{``Cifar task 1''}}
    \label{fig:task_cifar_indi}
    \end{subfigure}
    \hspace{0.03\textwidth}
    \begin{subfigure}[b]{0.3\linewidth}
        \includegraphics[width=1.0\textwidth]{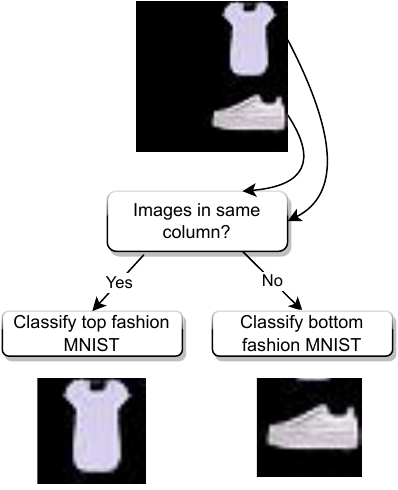}
    \caption{\textbf{``Top if above''}}
    \label{fig:task_top_if_above}
    \end{subfigure}
    \hspace{0.03\textwidth}
    \begin{subfigure}[b]{0.3\linewidth}
    \includegraphics[width=1.0\textwidth]{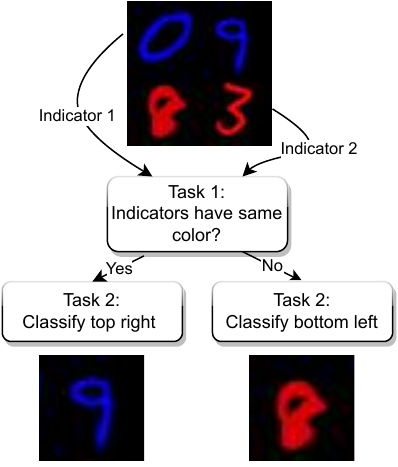}
    \caption{\textbf{``Same color decision''}}
    \label{fig:task_color_decision}
    \end{subfigure}
    
    \vspace{0.1\textwidth}
    \begin{subfigure}[b]{0.3\linewidth}
        \includegraphics[width=1.0\textwidth]{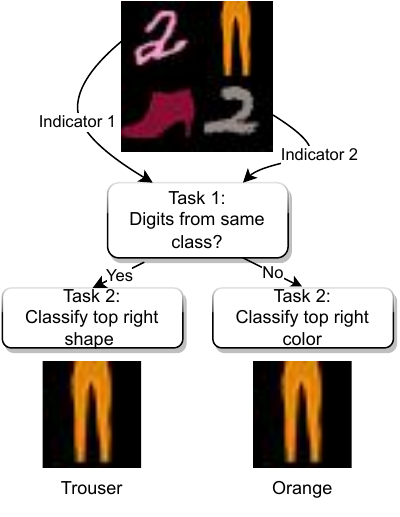}
    \caption{\textbf{```Color or fashion class''}}
    \label{fig:task_object_or_color}
    \end{subfigure}
    \hspace{0.2\textwidth}
    \begin{subfigure}[b]{0.3\linewidth}
    \includegraphics[width=1.0\textwidth]{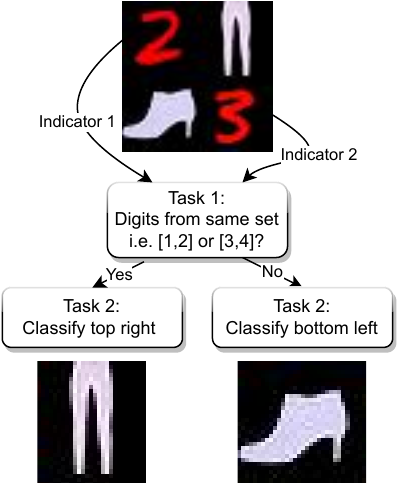}
    \caption{\textbf{``Digit grouping''}}
    \label{fig:digit_grouping}
    \end{subfigure}
\caption{\textbf{Schematics for additional datasets.} }
\label{fig:more_datasets}
\end{figure*}

In the following we report results on 5 more datasets. The datasets are depicted in Fig.~\ref{fig:more_datasets}.

\textbf{Cifar task 1.}
A schematic for this task is shown in Fig.~\ref{fig:task_cifar_indi}.
The ``Cifar task 1'' dataset uses Cifar-10 \cite{cifar} images of classes ``automobile'' and ``bird'' as indicators. 
Targets are sampled from fashion MNIST and MNIST. All 4 images are randomly placed on a 4x4 canvas and we apply random colors (red or blue) to the MNIST and fashion MNIST samples.
Task 1 is to compare the Cifar-10 classes.
If they come from the same class, task 2 is to classify the MNIST digit. If not, the tasks is to classify the fashion MNIST sample. 
Results are reported in Tab.~\ref{tab:cifar_task1}. normal ViT fails in 1/4 cases and \ahaepoch is usually late.
Note, that this task may seem difficult, but differences in color distribution of ``bird'' and ``automobile'' simplify the task.

\begin{table}[h]
    \vspace{-1\baselineskip}
  \caption{\textbf{Results ``Cifar task 1'' dataset}. $\tau$ not optimized for this task. \textbf{ER}: \aharatio, \textbf{Acc.}: Accuracy, \textbf{Avg.~EE.}: average \ahaepoch. }
  \label{tab:cifar_task1}
  \vspace{-1\baselineskip}
  \centering
    \resizebox{1.0\linewidth}{!}{
  \begin{tabular}{lc|rrr}
    \toprule
    & & \multicolumn{3}{c}{Cifar task 1}\\
    \midrule
    \multicolumn{2}{c}{} & \multicolumn{1}{|c}{} & \multicolumn{2}{c}{Avg.~over EMs}    \\
    \cmidrule(lr){4-5} 
    Model & $\tau$    & ER $\uparrow$    & Acc. $\uparrow$  & Avg.~EE. \\
    \midrule
    ViT & $\sqrt{d_k}$ &3/4 & 83.43 $\pm$ 1.83  &187.67 $\pm$ 42.46  \\
    ViT & $\frac{1}{3}$ &\textbf{4/4} & \textbf{86.86 $\pm$ 0.75}  & \textbf{76.50 $\pm$ 08.90}  \\
    \normsoftmax & $\sqrt{d_k}$ &\textbf{4/4} &82.89 $\pm$ 0.31  & 100.0 $\pm$ 27.89  \\
    \bottomrule
  \end{tabular}
  }
\end{table}

\textbf{Top if above.}
The task description is summarized in Fig.~\ref{fig:task_top_if_above}. 
For data creation we sample 2 images from fashion MNIST and place one in the top row of a 4x4 canvas and the other in the bottom row. The column is selected randomly for both. Task 1 is to check whether the 2 samples are in the same column. If they are, task 2 is to classify the top image. If not, the image in the bottom row must be classified.
This task is relatively simple, as it removes additional indicators. Instead, only the relative location of the images is the relevant information to solve task 1.
This task is very simple and leads to a low \ahaepoch for all methods (see \ref{fig:task_top_if_above}).
\begin{table}[h]
  \vspace{-1\baselineskip}
  \caption{\textbf{Results ``Top if above'' dataset}. $\tau$ not optimized for this task. \textbf{ER}: \aharatio, \textbf{Acc.}: Accuracy, \textbf{Avg.~EE.}: average \ahaepoch. }
  \label{tab:top_if_above}
  \vspace{-1\baselineskip}
  \centering
 \resizebox{1.0\linewidth}{!}{
  \begin{tabular}{lc|rrr}
    \toprule
    & & \multicolumn{3}{c}{Top if above}\\
    \midrule
    \multicolumn{2}{c}{} & \multicolumn{1}{|c}{} & \multicolumn{2}{c}{Avg.~over EMs}    \\
    \cmidrule(lr){4-5} 
    Model & $\tau$    & ER $\uparrow$    & Acc. $\uparrow$  & Avg.~EE. \\
    \midrule
    ViT & $\sqrt{d_k}$ &\textbf{4/4} & \textbf{91.47 $\pm$ 0.13} & 13.75 $\pm$ 2.19 \\
    ViT & $\frac{1}{3}$ &\textbf{4/4} & 90.38 $\pm$ 0.13 & 9.5 $\pm$ 1.25 \\
    \normsoftmax & $\sqrt{d_k}$ &\textbf{4/4} & 90.84 $\pm$ 0.27 & \textbf{9.25 $\pm$ 1.09} \\
    \bottomrule
  \end{tabular}
  }
\end{table}

\textbf{Same color decision task.}
The task is explained in Fig.~\ref{fig:task_top_if_above}. 
For data creation we sample only MNIST digits and apply random colors (red or blue) to all digits. If color of the indicators is identical, the top right must be classified and bottom left if not. 
As can be seen in Tab.~\ref{tab:same_color_decision_task}, this task is again very easy. Color seems to be easily accessible for ViT and ViT has little trouble to compare the indicator colors.
\begin{table}[h]
\vspace{-1\baselineskip}
  \caption{\textbf{Results ``Same color decision task'' dataset}. $\tau$ not optimized for this task. \textbf{ER}: \aharatio, \textbf{Acc.}: Accuracy, \textbf{Avg.~EE.}: average \ahaepoch. }
  \label{tab:same_color_decision_task}
  \vspace{-1\baselineskip}
  \centering
 \resizebox{1.0\linewidth}{!}{
  \begin{tabular}{lc|rrr}
    \toprule
    & & \multicolumn{3}{c}{Same color decision task}\\
    \midrule
    \multicolumn{2}{c}{} & \multicolumn{1}{|c}{} & \multicolumn{2}{c}{Avg.~over EMs}    \\
    \cmidrule(lr){4-5} 
    Model & $\tau$    & ER $\uparrow$    & Acc. $\uparrow$  & Avg.~EE. \\
    \midrule
    ViT & $\sqrt{d_k}$ & \textbf{4/4} &98.35 $\pm$ 0.58 & 91.5 $\pm$ 60.04 \\
    ViT & $\frac{1}{3}$ & \textbf{4/4} & \textbf{98.95 $\pm$ 0.10} & 8.25 $\pm$ 1.48 \\
    \normsoftmax & $\sqrt{d_k}$ & \textbf{4/4} & 98.93 $\pm$ 0.03 & \textbf{7.75 $\pm$ 0.43} \\
    \bottomrule
  \end{tabular}
  }
\end{table}

\textbf{Color or fashion classification.}
This task is shown in Fig.~\ref{fig:task_object_or_color}. For the creation of the dataset we define 10 random colors, i.e.,~(brown, blue, yellow, orange, red, green, purple, gray, pink, turquoise) and apply a random color to each target and each indicator sample. 
For targets we use fashion MNIST samples and indicators are MNIST classes ``1'' and ``2''. Task 1 is to compare digits. If they are the same class, the top right fashion sample must be classified. If not, the color of the top right sample must be classified.
\begin{table}
  \caption{\textbf{Results ``Color or fashion class'' dataset}. $\tau$ not optimized for this task. \textbf{ER}: \aharatio, \textbf{Acc.}: Accuracy, \textbf{Avg.~EE.}: average \ahaepoch. }
  \label{tab:color_or_fashion_class}
  \vspace{-1\baselineskip}
  \centering
   \resizebox{1.0\linewidth}{!}{
  \begin{tabular}{lc|rrr}
    \toprule
    & & \multicolumn{3}{c}{Color or fashion class}\\
    \midrule
    \multicolumn{2}{c}{} & \multicolumn{1}{|c}{} & \multicolumn{2}{c}{Avg.~over EMs}    \\
    \cmidrule(lr){4-5} 
    Model & $\tau$    & ER $\uparrow$    & Acc. $\uparrow$  & Avg.~EE. \\
    \midrule
    ViT & $\sqrt{d_k}$ &4/4 &92.85 $\pm$ 0.25 & 12.00 $\pm$ 3.32 \\
    ViT & $\frac{1}{3}$ &4/4 & 92.53 $\pm$ 0.40 & 20.75 $\pm$ 2.49 \\
    \normsoftmax & $\sqrt{d_k}$ &4/4 & \textbf{92.75 $\pm$ 0.71} & \textbf{11.25 $\pm$ 0.83} \\
    \bottomrule
  \end{tabular}
}
\end{table}

\textbf{Digit grouping.}
Finally, we make the indicator task more difficult. We follow the same setting as for the \emph{Main dataset}, as described in Fig.~\ref{fig:teaser_data}. However, indicators are not sampled from digits 1 and 2, but from 1, 2, 3 and 4. Task 1 is to find out whether both indicators are smaller or both indicators are larger or equal to 3. I.e., we build indicator sets $[1,2]$ and $[3,4]$ if both indicators are from the same group the top-right image should be classified.
As can be seen in Tab.~\ref{tab:digit_grouping}, increasing the difficulty of the indicator task quickly makes the dataset too hard. Further optimization of hyper-parameters and architecture are likely to solve the tasks. 
\begin{table}
  \caption{\textbf{Results ``Digit grouping'' dataset}. $\tau$ not optimized for this task. \textbf{ER}: \aharatio, \textbf{Acc.}: Accuracy, \textbf{Avg.~EE.}: average \ahaepoch. }
  \label{tab:digit_grouping}
  \vspace{-1\baselineskip}
  \centering
     \resizebox{1.0\linewidth}{!}{
  \begin{tabular}{lc|rrr}
    \toprule
    & & \multicolumn{3}{c}{Digit grouping}\\
    \midrule
    \multicolumn{2}{c}{} & \multicolumn{1}{|c}{} & \multicolumn{2}{c}{Avg.~over EMs}    \\
    \cmidrule(lr){4-5} 
    Model & $\tau$    & ER $\uparrow$    & Acc. $\uparrow$  & Avg.~EE. \\
    \midrule
    ViT & $\sqrt{d_k}$ &0/4 & -  & - \\
    ViT & $\frac{1}{3}$ &0/4 & -  & - \\
    \normsoftmax & $\sqrt{d_k}$ &0/4 & - & - \\
    \bottomrule
  \end{tabular}
}
\end{table}

\subsubsection{What makes a multi-step task hard to learn?}
Based on the additional datasets described above we observe, that task 1 difficulty plays a major role in what makes a multi-step task difficult. For example the \emph{Same color decision task} (\cref{fig:task_color_decision} and \cref{tab:same_color_decision_task}) is significantly easier than our \emph{Main task} (compare to Avg. \ahaepoch in \cref{tab:final_res_joint_with_std}). We attribute this to the readily available feature (color) used to solve task 1. Similarly, the \emph{Top if above} task (\cref{fig:task_top_if_above} and \cref{tab:top_if_above}) is very easy to learn, as it removes one task 1 feature completely and relies solely on the position features.
Making task 1 harder leads to much rarer and later \ahas, as can be seen for the \emph{No position task} (\cref{fig:tas} and \cref{tab:final_res_no_pos_with_std}) and the \emph{Digit grouping task} (\cref{fig:digit_grouping} and \cref{tab:digit_grouping}). 
They make task 1 harder by removing the position information to locate and match indicators or introducing more variance to the appearance of indicators, respectively.
We conjecture, that task 1 difficulty plays a major role in determining whether and when a \aha will happen. Task difficulty here seems to depend on the availability of the features required to solve task 1.

\subsection{\ahas on Realistic, Large Scale High-resolution Images (ImageNet-100 based)}
\label{sec:imagenet}

\begin{figure*}[t]
    \centering
    \begin{subfigure}[l]{0.4\textwidth}
        \includegraphics[width=1\textwidth,trim=8 5 0 3,clip]{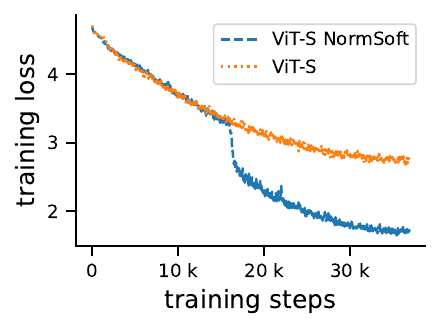}
        \vspace{-1.5\baselineskip}
        \caption{Training loss on ImageNet-100 dog decision for ViT-S trained with \normsoftmax and vanilla ViT-s}
        \label{fig:imgnet_aha_loss}
    \end{subfigure}
    \hspace{0.075\textwidth}
    \begin{subfigure}[r]{0.4\textwidth}
        \includegraphics[width=1\textwidth,trim=8 5 0 3,clip]{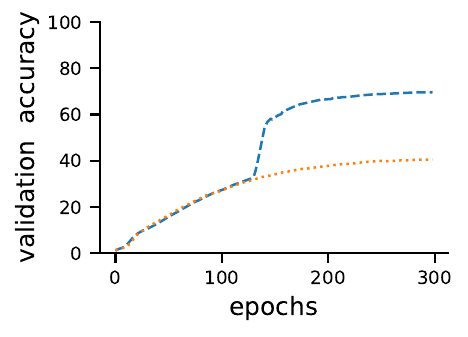}
        \vspace{-1.5\baselineskip}
        \caption{Validation accuracy on ImageNet-100 dog decision for ViT-S trained with \normsoftmax and vanilla ViT-s}
        \label{fig:imgnet_aha_acc}
    \end{subfigure}
    \caption{Results on ImageNet-100 based task. Also for a larger ViT, realistic data, high-resolution images we observe \ahas. In particular, \normsoftmax leads to a \aha, while the vanilla ViT fails to learn task 1. Task description: If both indicators show the image of an identical dog the top-right image is the target and bottom left otherwise. Dog samples taken from 2 dog classes. Probability of top-right being the target is set to 0.5.}
    \label{fig:imagenet_aha}
\end{figure*}

To show that \ahas can also be observed on large scale high-resolution datasets we create a dataset with high-resolution natural images using ImageNet-100 \cite{tian2020contrastive}. For simplicity, we follow the same dataset design as for the MNIST-like datasets, i.e.,~we place the targets in the top-right and bottom left, while the other two quadrants show indicator images. Targets are simply images from ImageNet-100. The indicators are sampled from 2 of the ImageNet dog classes. If both indicators show the exact same sample, the top-right image needs to be classified and bottom left otherwise. The probability of top-right location being the target is 0.5. We train ``ViT-S" and ``ViT-S with \normsoftmax" following the standard ImageNet training setting using strong augmentations from Deit training \cite{deit}. Vanilla ViT obtains only ~41\% accuracy and we don't observe a Eureka-Moment. For ViT-S with NormSoftmax we observe a Eureka-Moment at epoch 131. The training and validation curves are shown in Fig.~\ref{fig:imagenet_aha}.

\subsection{Vanishing Gradient in the Softmax.}
\label{sec:vanishing_gradient}

\textbf{Softmax attention can cause vanishing gradients for $W_q$ and $W_K$.}
To see that softmax attention can result in vanishing gradients it helps to take a look at the gradients of the attention function. 
Let
\begin{equation}
    A(W_q, W_k, W_v, X) = Z 
\end{equation}
\begin{multline}
 A(W_q, W_k, W_v, X) = \\ S\Bigl(D\bigl(Q(W_q, X), K(W_k, X)\bigr)\Bigr)(W_v, X), 
\end{multline}
\begin{align}
    S(D) &= \text{softmax}(D), \\
    D(Q, K) &= \frac{QK^T}{\tau}, \\
    Q(W_q, X) &= W_q X, \\
    K(W_k, X) &= W_k X, \\
    V(W_v, X) &= W_v X.
\end{align}
be the attention function, where $W_k$, $W_q$ and $W_v$ are weight matrices, $X$ is the input.

Using the chain rule we get
\begin{align}
    \frac{\partial A}{\partial W_q} =& \frac{\partial A}{\partial D} \frac{\partial D}{\partial Q} \frac{\partial Q}{\partial W_q}\\
    \frac{\partial A}{\partial W_k}=& \frac{\partial A}{\partial D} \frac{\partial D}{\partial K} \frac{\partial K}{\partial W_k}.
\end{align}

Since  $\frac{\partial D}{\partial Q}, \frac{\partial Q}{\partial W_q}, \frac{\partial D}{\partial K}, \frac{\partial K}{\partial W_k}$ are constants we only need to look more closely into $ \frac{\partial A}{\partial D}$.

$\frac{\partial A}{\partial D}$ is given by $\frac{\partial A}{\partial D}=\frac{\partial S}{\partial D}V$, where $S(D)$ takes the values $S=(s_1,\dots,s_n)$. 
Therefore, to analyze how the gradients  $\frac{\partial A}{\partial W_q}$ and $\frac{\partial A}{\partial W_k}$ behave, we need to analyze the $\frac{\partial S}{\partial D}$, i.e.,~the Jacobian of the Softmax $S(D)$.
It is given by
\begin{align}
    \frac{\partial S}{\partial D} = 
    \begin{pmatrix}
    s_1 (1-s_1) & -s_1 s_2  & \dots & -s_1 s_n \\
    -s_2 s_1 & s_2 (1-s_2) & \dots & -s_2 s_n \\
    \vdots & \vdots & \ddots & \vdots \\
    -s_n s_1 & -s_n s_2 & \dots & s_n (1-s_n)
    \end{pmatrix}.
\end{align}
It can be easily seen, that almost all entries in the Jacobian are close to 0 whenever a single $s_i$ is close to 1 and all others are almost 0. Please see \cite{kurbiel2021} for a more detailed analysis.

\subsection{Information-Theoretic Probes}
\label{sec:info_probes}
While our main analysis of accessibility of information in the main paper is entirely sufficient to support our claims, a more detailed look into accessibility of the information can complement our analysis.
Our analysis tests for linear decodability of particular features from the learned representations. However, it does not measure how easily information is accessible. \citet{voita2020information} propose to measure this ``effort'' by changing the amount of data needed to learn to extract the feature in question, referred to as online codelength. The codelength is measured in bits.
The idea of online codelength is to measure the ``availability'' of a feature by limiting the data to learn the probe. 
Intuitively, if the representation has a high degree of order and the feature in question can be easily separated from other features, less data is required to learn a good representation. 
Thus, a short codelength corresponds to a feature being easily accessible.
Following \citet{voita2020information}, we use $0.1\%$, $0.2\%$, $0.4\%$, $0,8\%$, $1.6\%$, $3.2\%$, $6.25\%$, $12,5\%$, $25\%$ and $50\%$ of the data to learn the probes.
We train each probe for 70 epochs, e.g., for the first subset ($0.1\%$) the linear probe sees the same $0.1\%$ of the dataset 70 times.

\begin{figure*}
    \centering
    \begin{subfigure}[c]{1\linewidth}
        \includegraphics[width=1\linewidth]{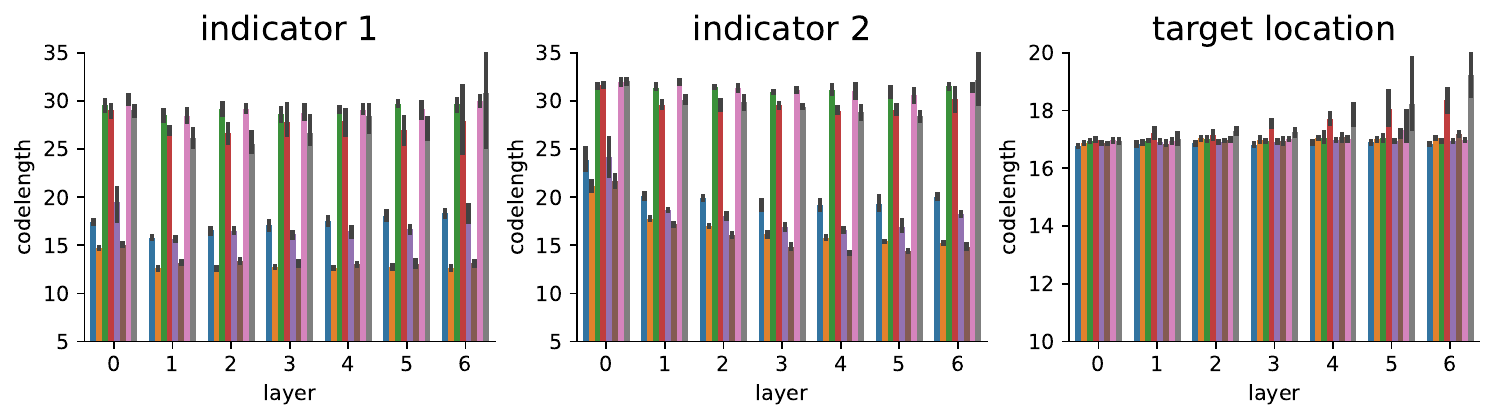}
        \caption{ViT (without \aha)}
        \label{fig:info_probe_vit}
    \end{subfigure}
        \begin{subfigure}[c]{1\linewidth}
        \includegraphics[width=1\linewidth]{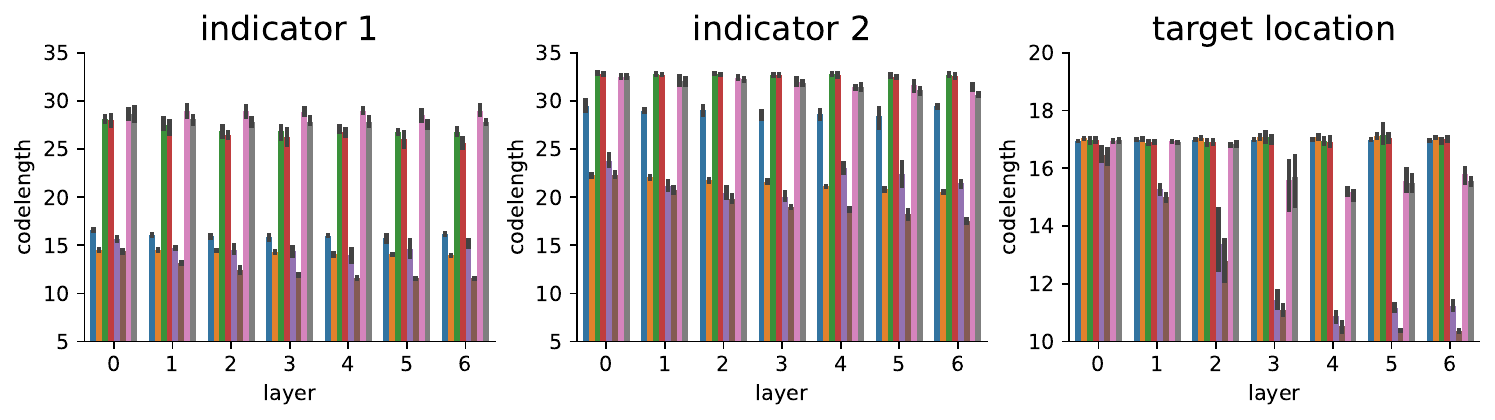}
        \caption{ViT+HT (with \aha)}
        \label{fig:info_probe_ht}
    \end{subfigure}
    \begin{subfigure}[c]{1\linewidth}
        \includegraphics[width=1\linewidth]{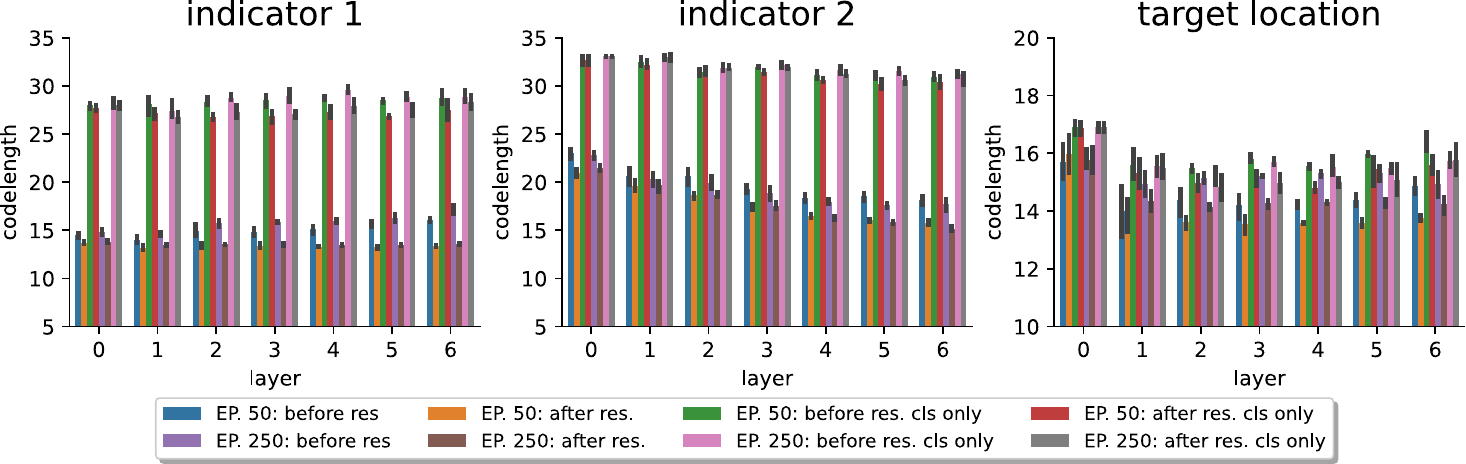}
        \caption{ViT $\tau=\frac{1}{3}$ (with \aha)}
        \label{fig:info_probe_constant}
    \end{subfigure}
    \caption{\textbf{How easily is information accessible in different parts of the network for} \textbf{(a)} ViT without \aha, \textbf{(b)} ViT+HT (with \aha) and \textbf{(c)} ViT $\tau=\frac{1}{3}$ (with \aha). To measure how easily information can be extraceted we follow the online codelength approach of \citet{voita2020information}. In contrast to \cref{fig:linprobe_vit_failure}, lower is better for codelength.}
    \label{fig:info_probe}
\end{figure*}

As can be seen in \cref{fig:info_probe}, the codelength necessary to extract the indicators does not differ much. 
Similarly as in \cref{fig:linprobe_vit_failure}, differences can be observed for the target location.
For ViT (\cref{fig:info_probe_vit}) we see that the codelength for the target location is large for all types of representations across all layers. For ViT+HT (\cref{fig:info_probe_ht}) we can see that late in training the feature becomes easily accessible, in particular in deeper layers and in the full representation. 
In contrast, ViT $\tau=\frac{1}{3}$ (\cref{fig:info_probe_constant}) represents the feature already early in training and also the CLS token contains a better representation of the target location compared to ViT+HT.

\subsection{Linear Probe Results for $Q$, $K$ and $V$ and for Target Classification.}

In the following we will show the linear probe results for  $Q$, $K$, $V$ and $Z$ for all layers.

\begin{figure*}[h]
    \centering
    \includegraphics[width=1.0\textwidth]{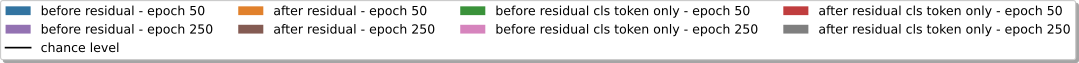}
    \begin{subfigure}[b]{0.49\linewidth}
   \includegraphics[width=1.0\textwidth]{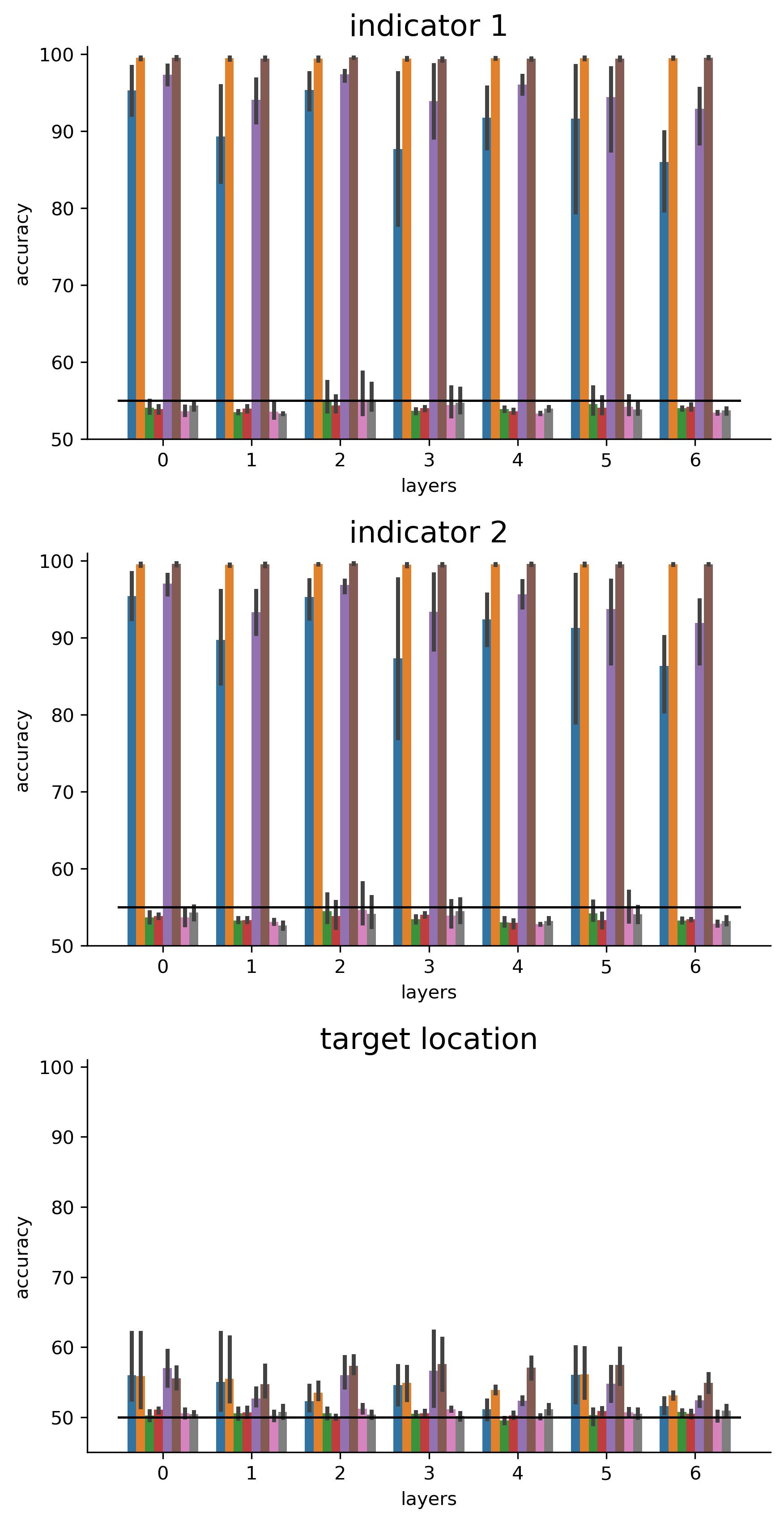}
   \caption{ViT without \aha}
   \label{fig:linprobe_z_no_aha}
   \end{subfigure}
       \begin{subfigure}[b]{0.49\linewidth}
   \includegraphics[width=1.0\textwidth]{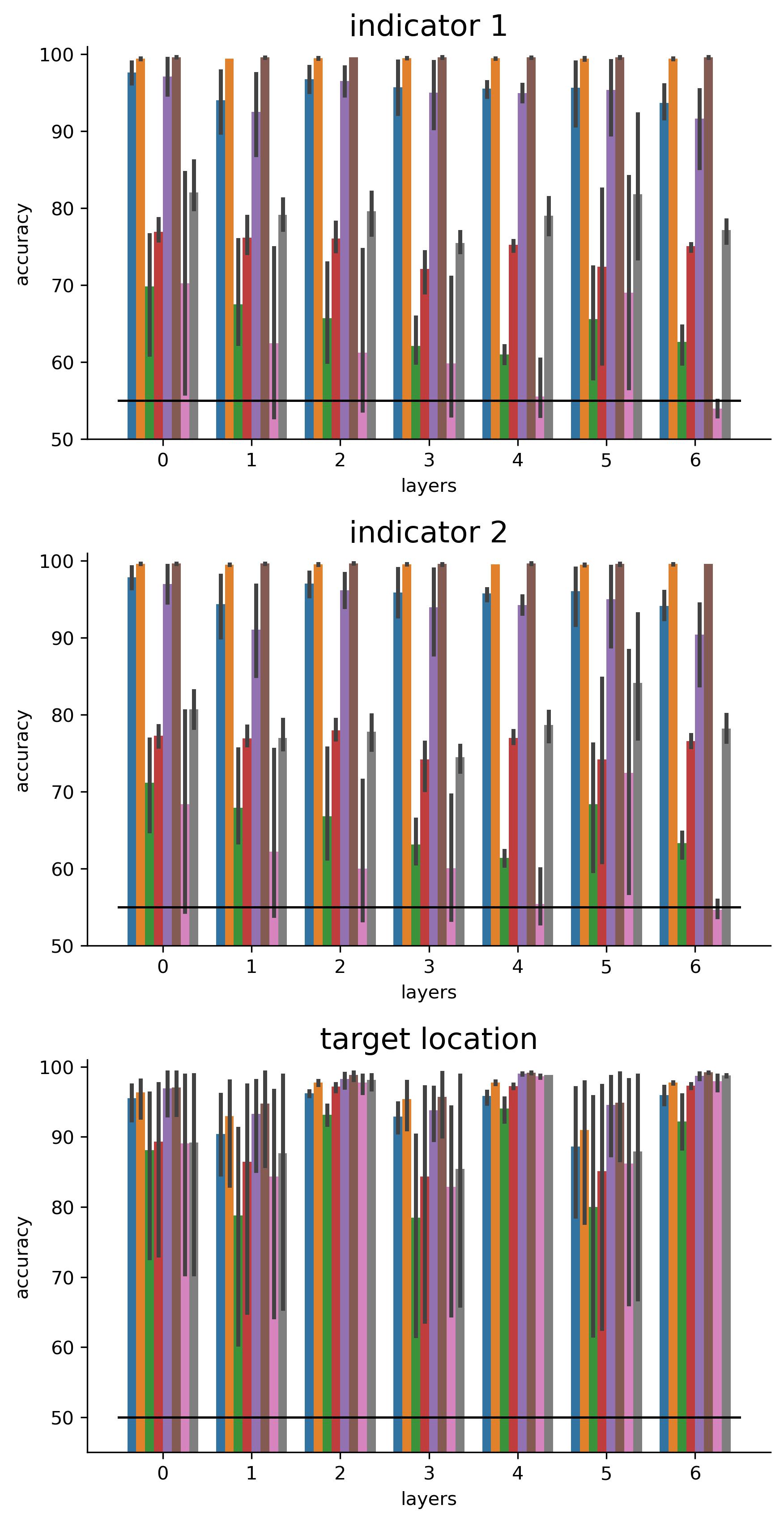}
   \caption{ViT with \aha}
   \label{fig:linprobe_z_with_aha}
   \end{subfigure}
       \caption{\textbf{Linear probe results for with $Z$ as input (all layers).} This the complete version of Fig.~\ref{fig:linprobe_vit_failure}. Additionally, we can observe in (b) that layers 2, 3 and 6 contain more target location information than other layers, indicating, that this information can be processed in these layers.}
    \label{fig:linprobe_z}
\end{figure*}

\textbf{Linear probe results for $Z$ for all layers.}
Fig.~\ref{fig:linprobe_z} shows the same plot as in the main paper, but for all layers. 
In addition to the observations made for the main paper, we can see that layer 2, 4 and 6 for the ViT with \aha (Fig.~\ref{fig:linprobe_z_with_aha}) represent significantly more information about the target locations than other layers in the CLS token. Indicating, that this information is extracted in these layers and written on the CLS token.

\begin{figure*}
    \centering
    \includegraphics[width=1.0\textwidth]{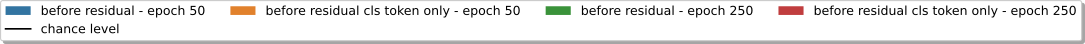}
    \begin{subfigure}[b]{0.49\linewidth}
   \includegraphics[width=1.0\textwidth]{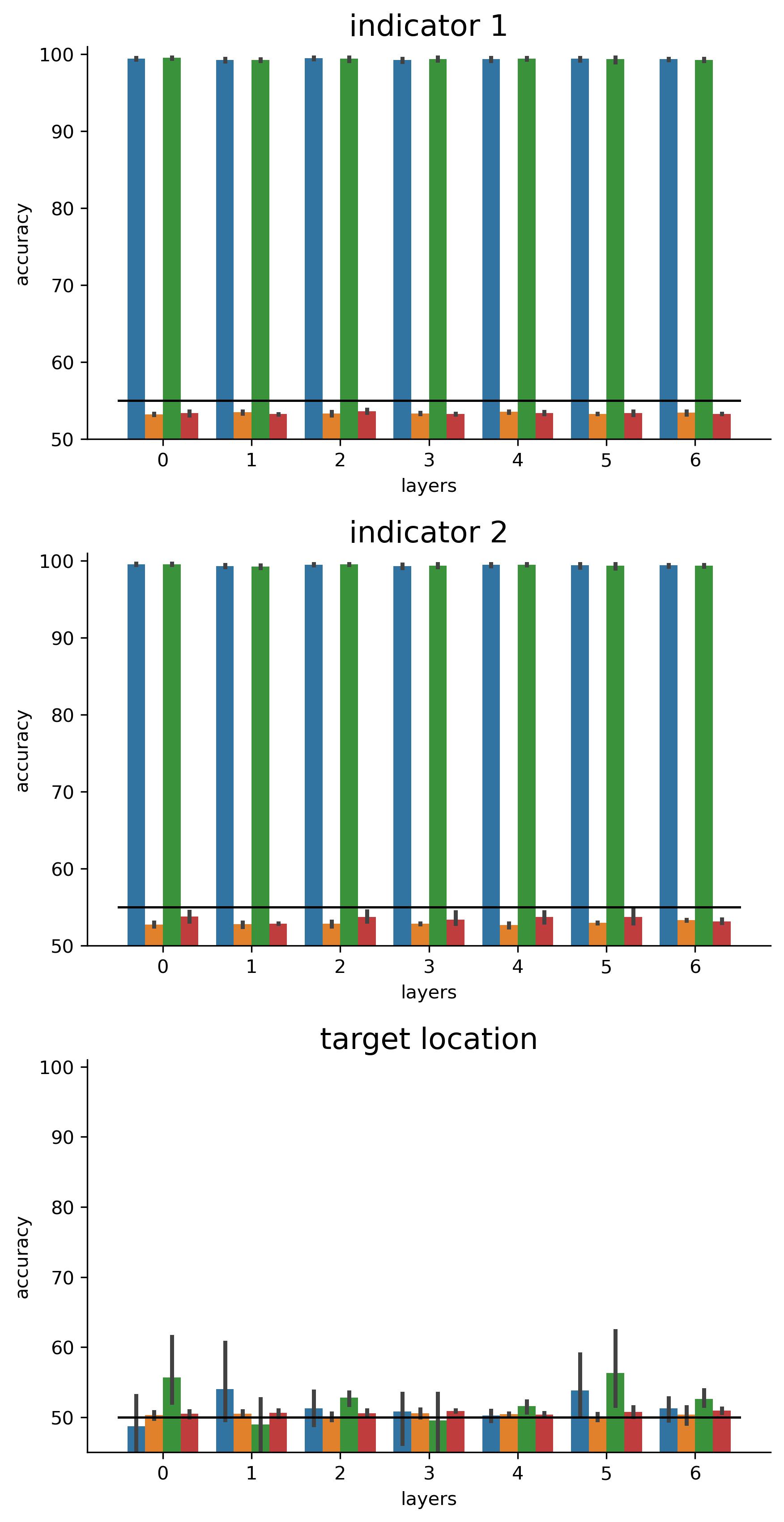}
   \caption{ViT without \aha}
   \label{fig:fig:linprobe_q_no_aha}
   \end{subfigure}
       \begin{subfigure}[b]{0.49\linewidth}
   \includegraphics[width=1.0\textwidth]{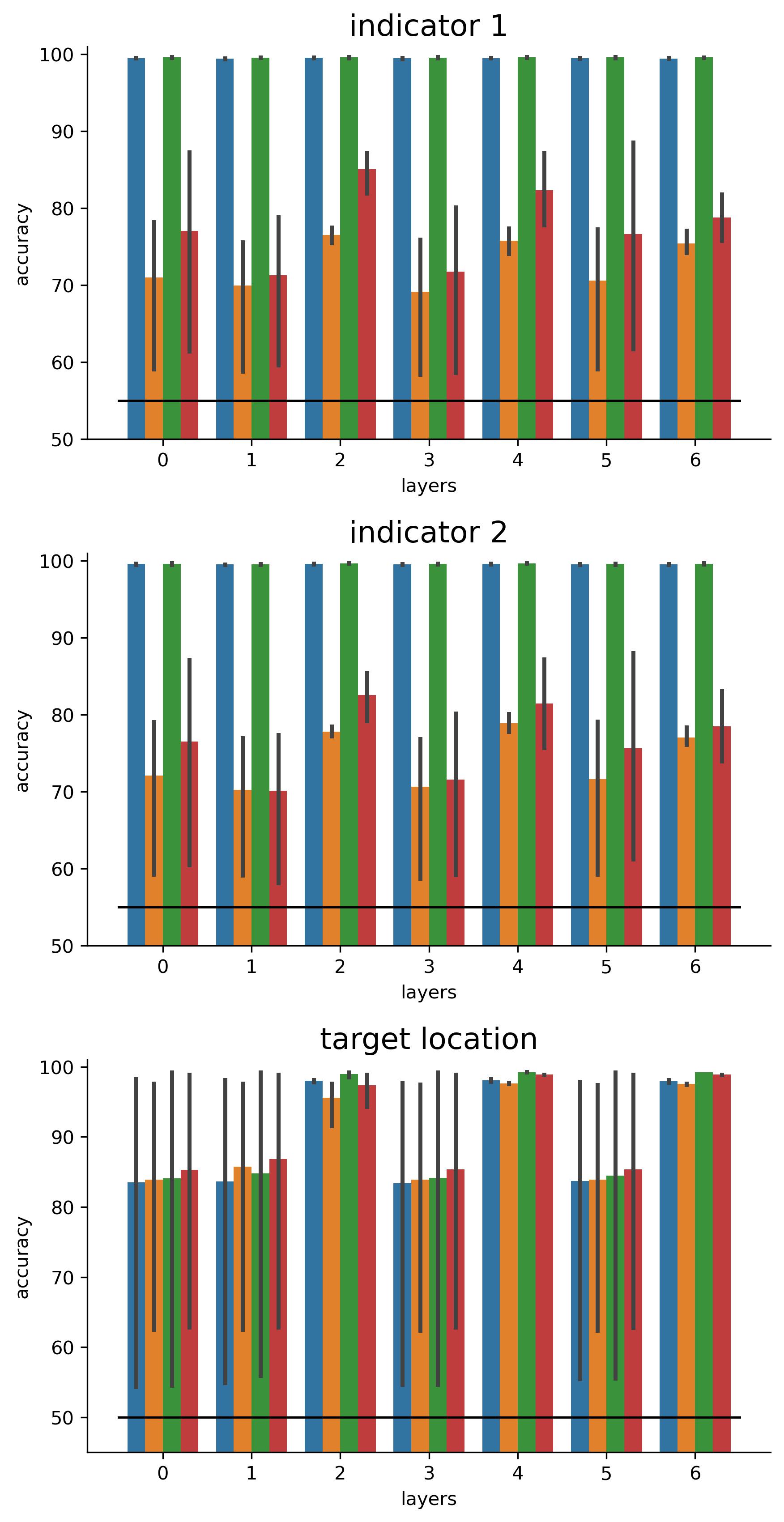}
   \caption{ViT with \aha}
   \label{fig:linprobe_q_with_aha}
   \end{subfigure}
       \caption{\textbf{Linear probes for Q.}}
    \label{fig:linprobe_q}
\end{figure*}

\begin{figure*}
    \centering
    \includegraphics[width=1.0\textwidth]{Figures/axes_qkv.png}
    \begin{subfigure}[b]{0.49\linewidth}
   \includegraphics[width=1.0\textwidth]{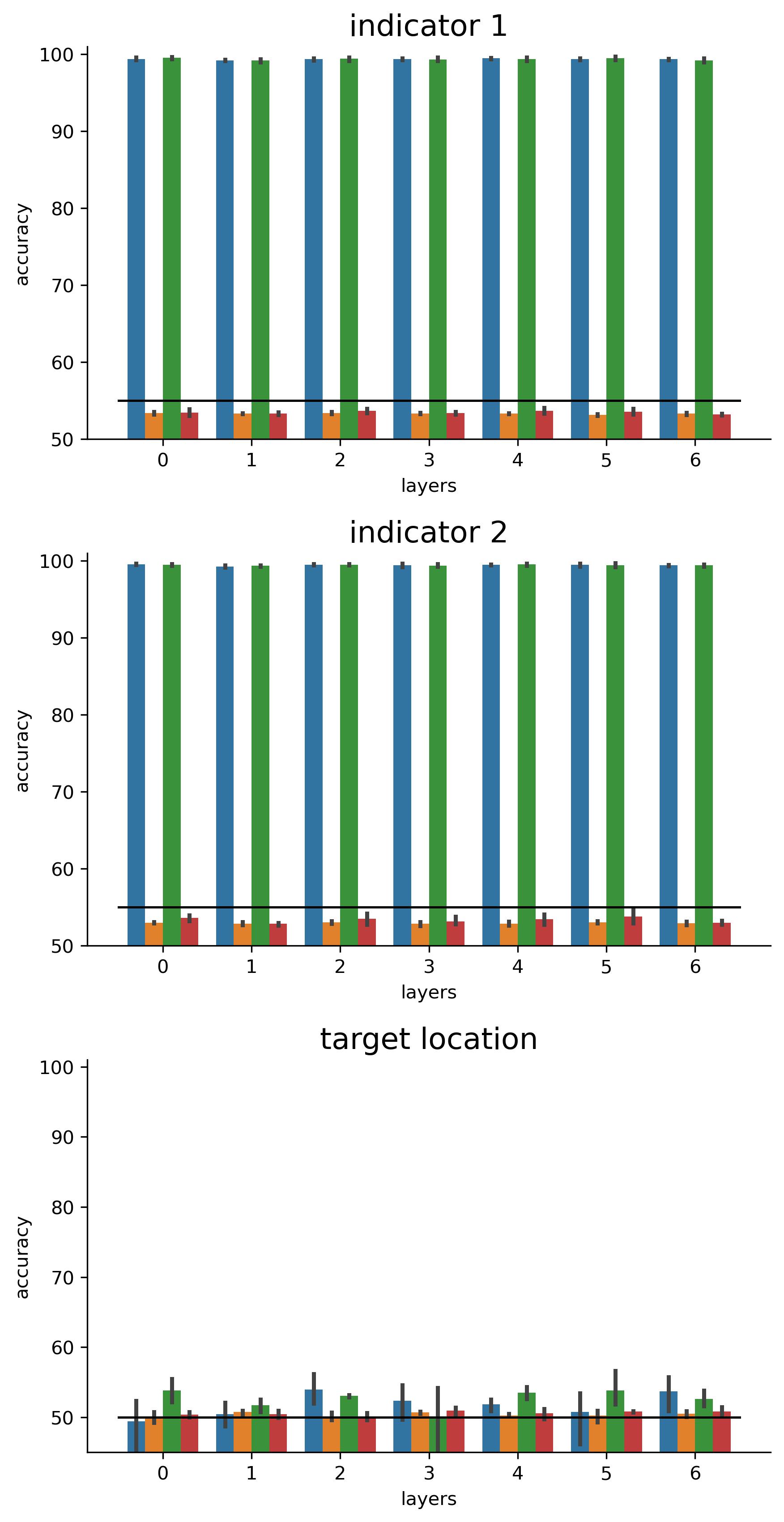}
   \caption{ViT without \aha}
   \label{fig:linprobe_k_no_aha}
   \end{subfigure}
       \begin{subfigure}[b]{0.49\linewidth}
   \includegraphics[width=1.0\textwidth]{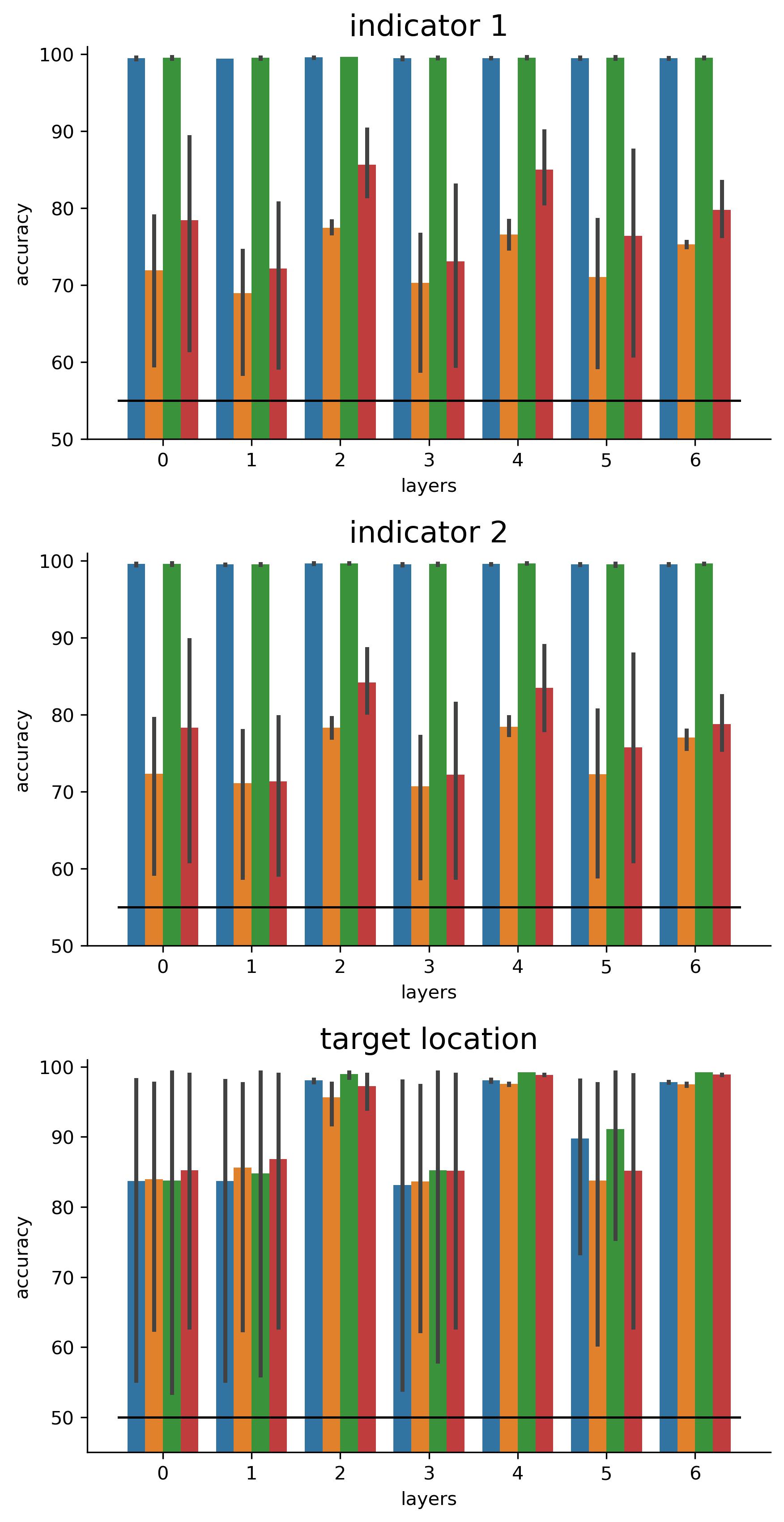}
   \caption{ViT with \aha}
   \label{fig:linprobe_k_with_aha}
   \end{subfigure}
       \caption{\textbf{Linear probes for K.}}
    \label{fig:linprobe_k}
\end{figure*}

\textbf{Linear probes for $Q$ and $K$ and $V$.}
Figs.~\ref{fig:linprobe_q} \& \ref{fig:linprobe_k} show the results when using the $Q$ and $K$ as input for the linear probes. Note, that $Q$ and $K$ are not updated by the residual connection of the attention block, therefore, no bars for ``after residual'' are plotted.
The linear probe classification accuracies for $Q$ and $K$ are very similar. 
Again, we can see that the ViT without \aha does not represent indicator information in the CLS token and target location can not be linearly separated from other information.
Similarly, as for linear probe results with $Z$, layer 2, 4 and 6 for $Q$ and $K$ contain significantly more information about the indicators and target location for the ViT with \aha (compare  Figs.~\ref{fig:linprobe_z_with_aha}, \ref{fig:linprobe_q_with_aha}, \ref{fig:linprobe_k_with_aha}).

\begin{figure*}
    \centering
    \includegraphics[width=1.0\textwidth]{Figures/axes_qkv.png}
    \begin{subfigure}[b]{0.49\linewidth}
   \includegraphics[width=1.0\textwidth]{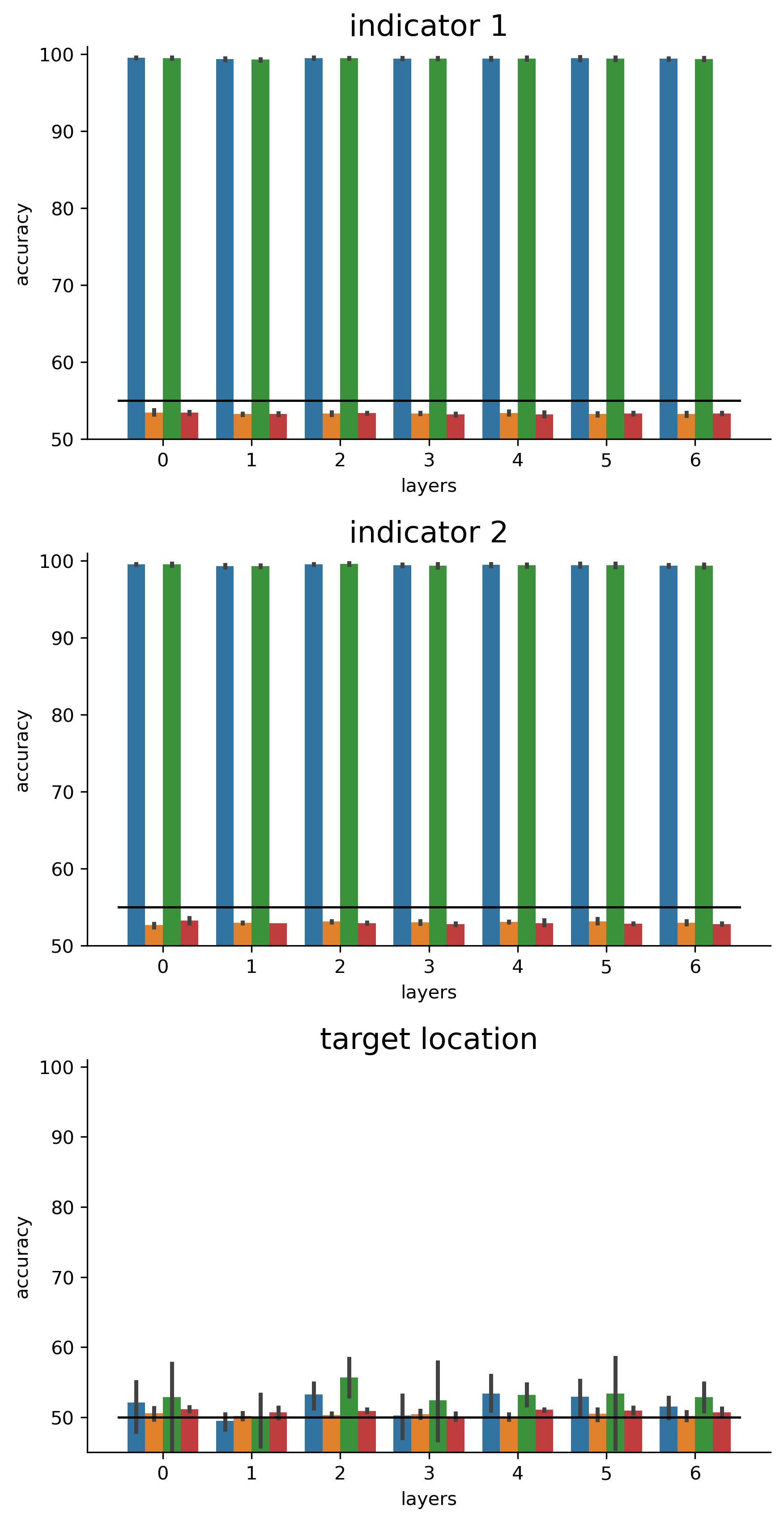}
   \caption{ViT without \aha}
   \label{fig:linprobe_v_no_aha}
   \end{subfigure}
       \begin{subfigure}[b]{0.49\linewidth}
   \includegraphics[width=1.0\textwidth]{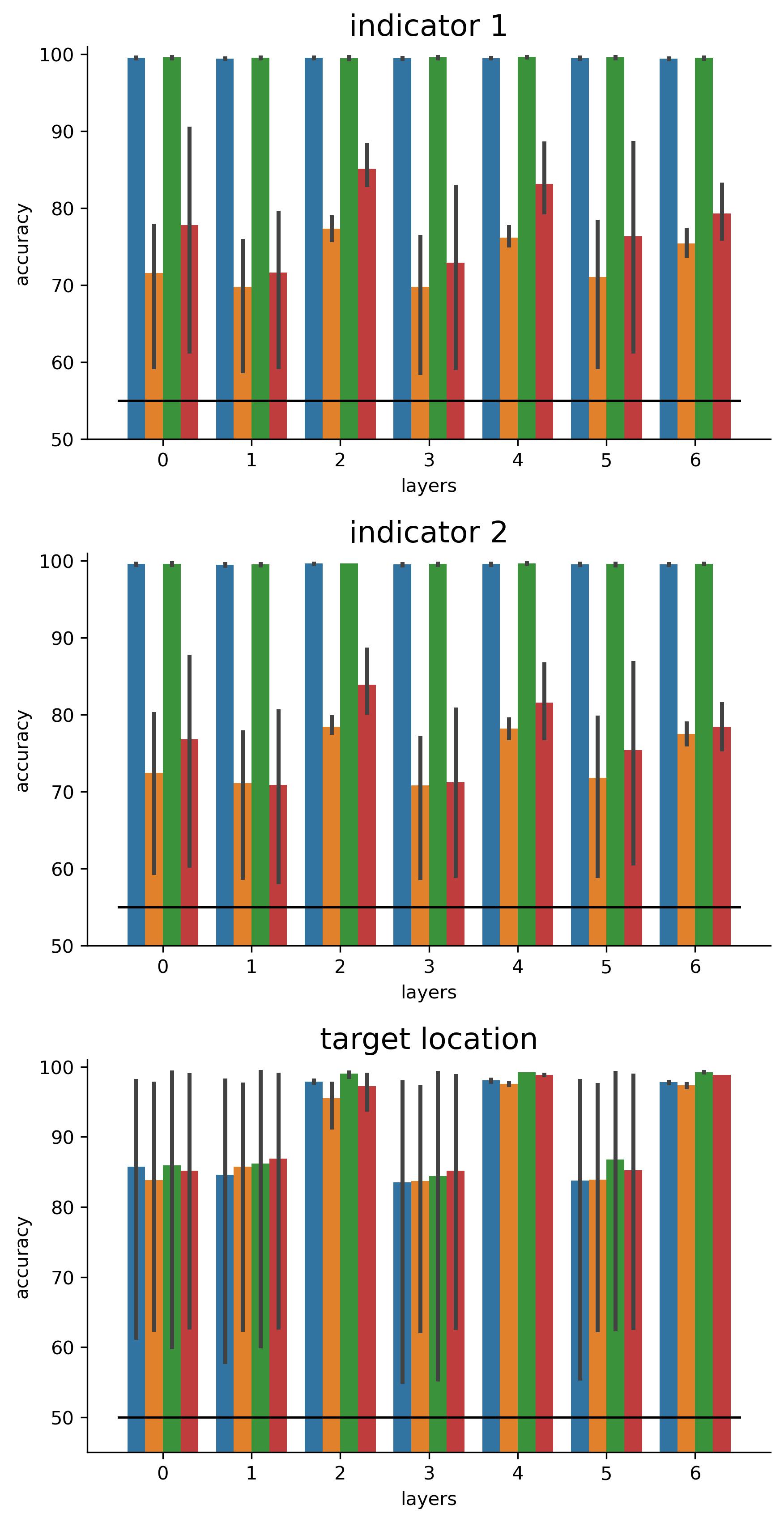}
   \caption{ViT with \aha}
   \label{fig:linprobe_v_with_aha}
   \end{subfigure}
       \caption{\textbf{Linear probes for V.}}
    \label{fig:linprobe_v}
\end{figure*}

The linear probe results for $V$ look very similar to those for $Q$ and $K$ (see Fig.~\ref{fig:linprobe_v}).

\begin{figure*}
    \centering
    \includegraphics[width=1.0\textwidth]{Figures/axes_z.png}
    \begin{subfigure}[b]{0.49\linewidth}
   \includegraphics[width=1.0\textwidth]{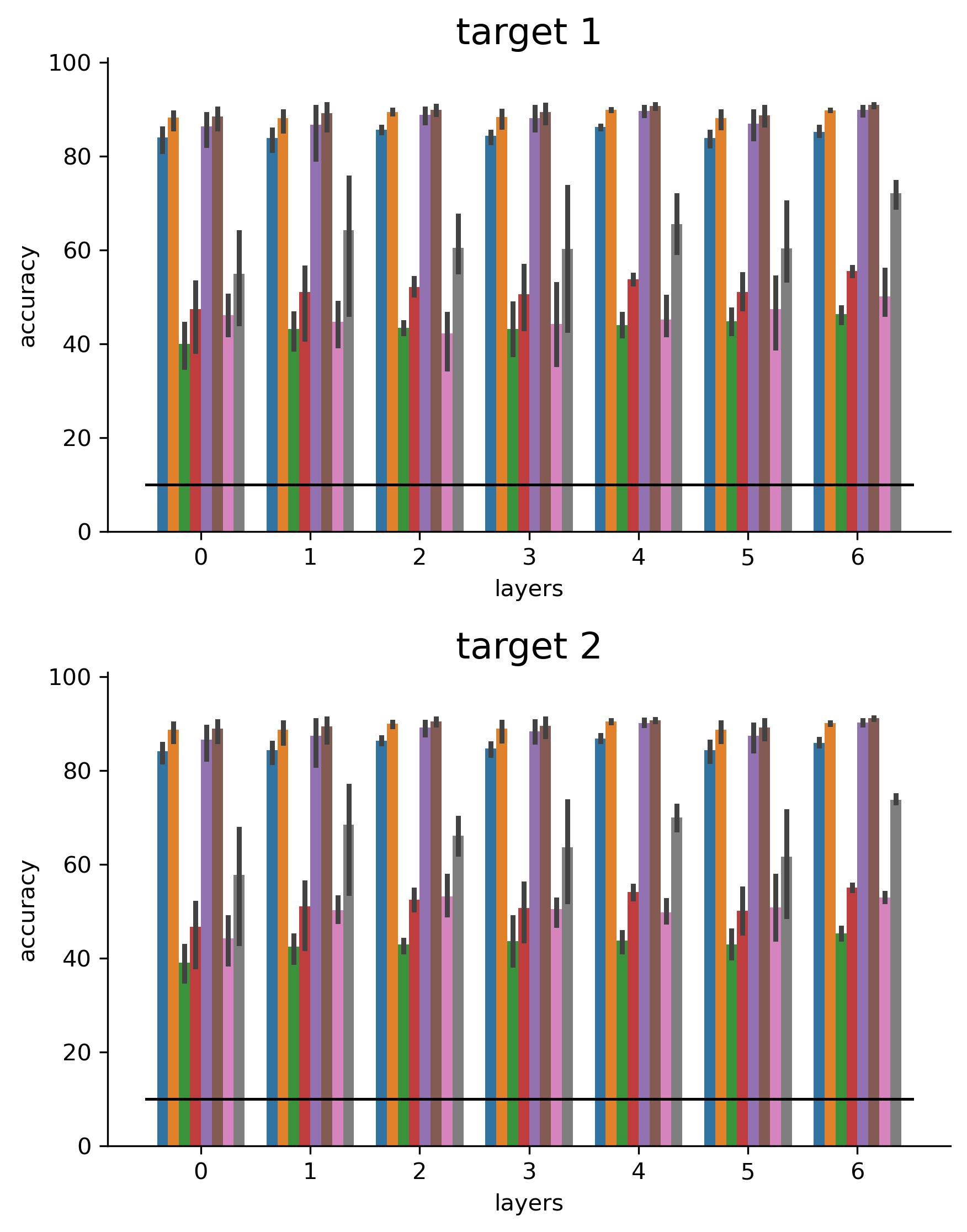}
   \caption{ViT without \aha}
   \label{fig:linprobe_target_z_no_aha}
   \end{subfigure}
       \begin{subfigure}[b]{0.49\linewidth}
   \includegraphics[width=1.0\textwidth]{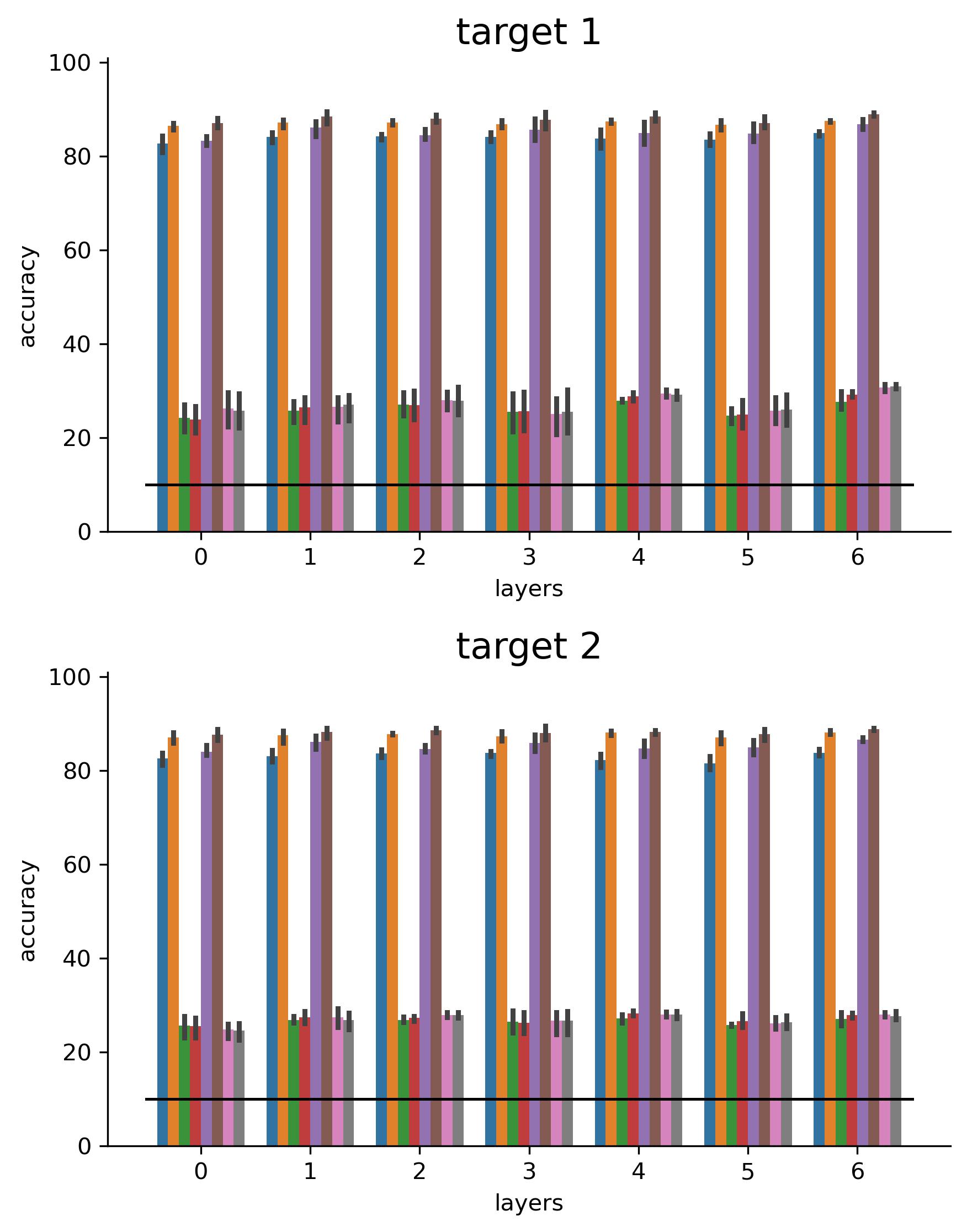}
   \caption{ViT with \aha}
   \label{fig:linprobe_target_z_with_aha}
   \end{subfigure}
       \caption{\textbf{Linear probes for Z with target classification.}}
    \label{fig:linprobe_target_z}
\end{figure*}

\textbf{Linear probes from $Z$ to targets.}
Last, we show the linear probe results when predicting the targets from $Z$. As can be seen, from the entire representation, for both ViT with \aha and ViT without \aha, target classes can be predicted with high  accuracy. Differences can be observed when using only the CLS token. Here, we observe that more target information is in the CLS token of the model without \aha (see Fig.~\ref{fig:linprobe_target_z}).

\subsection{Experimental setup --- Vision Task}
\label{sec:experimental_setup_vision}
We mostly follow the DeiT \citep{deit} training recipe without distillation.
For optimization we use AdamW \citep{loshchilov2018decoupled} with default values, i.e.,~$\beta_1=0.9$ and $\beta_2=0.999$ and $\epsilon=10^{-8}$.
Unless otherwise stated we Warmup the learning rate for 5 epochs from $10^{-6}$ to the maximum learning rate, use a weight decay of 0.05 and train for 300 epochs. 
We train all models with a batch size of 512, which fits on a single V100, for all the architectures that we considered.
Since position and color can be important cues in our datasets we train without data augmentation. We only sample color-noise from a standard normal Gaussian with standard deviation 0.05 for each color channel independently. Color-noise is sampled for each sub-sample (i.e.,~each indicator and each target) independently and added to the RGB value.

\textbf{Learning rate schedules for all models.}
To compare the tested methods and models fairly, we run a search over 9 learning rate schedules with 4 random seeds, each. 
We anneal the learning rate from a maximum to a minimum using a cosine scheduler. We also use Warmup, as described in the paragraph above. The different schedules can be seen in Tab.~\ref{tab:lr_schedule}.
We pick the schedule that leads to highest \aharatio for each model. In case of a tie we pick the schedule with higher accuracy.
\begin{table}[h]
    \centering
    \caption{\textbf{Learning rate schedules.} We use cosine annealing from ``max learning rate'' to ``min learning rate''.}
    \label{tab:lr_schedule}
    \begin{tabular}{r|r}
        \toprule
        max learning rate & min learning rate \\
        \midrule
         $10^{-3}$ & $10^{-5}$ \\
         $10^{-3}$ & $5* 10^{-6}$ \\
         $10^{-4}$ & $10^{-5}$ \\
         $5*10^{-4}$ & $5 * 10^{-6}$ \\
         $5*10^{-4}$ & $10^{-6}$ \\
         $10^{-4}$ & $10^{-6}$ \\
         $5*10^{-5}$ & $10^{-6}$ \\
         $10^{-5}$ & $10^{-6}$ \\
         $10^{-5}$& $10^{-7}$ \\
         \bottomrule
    \end{tabular}

\end{table}

\textbf{Info on taus:}
In initial experiments we tested for ViT  5 values for $\tau$, $\tau=\frac{2}{3}$, $\tau=\frac{1}{2}$, $\tau=\frac{1}{3}$, $\tau=\frac{1}{4}$ and $\tau=\frac{1}{5}$. We found $\tau=\frac{1}{3}$, $\tau=\frac{1}{4}$ to work well and did not further optimize them for the different methods. 
For HT we set the goal temperature to the default $\sqrt{d_k}$ and tried also  $\frac{1}{2*\sqrt{d_k}}$. Further optimizing these parameters for each model and dataset will most likely lead to improvements, but would add very little to a deeper understanding of the \ahas.

\subsection{Implementation Details \normsoftmax}

In practice, $\sigma(\cdot)$ can be defined by arbitrary functions. As highlighted in the background section, the standard deviation is a theoretically motivated choice. Alternatives are discussed by \citet{normsoftmax}. In this work, we find the variance to work better for ViT and RoBERTa, while we stick to the standard deviation for the reasoning task.

\subsection{Experimental Setup -- reasoning task}
\label{sec:reasoning_task_suppmat}
The input of the model is of the form ``a b c d =``, where, where $a,b,c,d \in \{0, 1, \dots, n\}$.
In our experiments, we set \begin{math}n=11\end{math}.
We train the transformer on 30\% of the entire set of possible inputs (i.e., \begin{math}11^4=14\,641\end{math} input combinations), that is with a batch size of 4392. 
The rest is used as test set.
We train for 10\,000 epochs over five random seeds.
We use token embeddings of size of \begin{math}d=2^{\lceil \log_2 n \rceil}=16\end{math}, four attention heads of dimension of \begin{math}d/4=4\end{math}, \begin{math}4d=64\end{math} hidden units in the MLP, and learned positional embeddings.

We trained with full batch gradient descent using AdamW \citep{loshchilov2018decoupled} with a cross-entropy loss. We optimized learning rates via grid search over \begin{math}[10^{-4},\;10^{-2}]\end{math} on seed 0. Following \citet{mechanistic_grokking} we use a weight decay of 1.

\subsection{Slingshot Effects on Reasoning Task} 
We observed that \normsoftmax caused slingshot effects \citep{thilak2022the} during the convergence phase of some of the training runs but believe this may be due to the interaction of gradients at different scales with adaptive optimizers \citep{mechanistic_grokking}. Since slingshot effects only occur after \ahas, they cannot be the cause for their occurrence.
We did not further investigate this observation.

\subsection{Experimental Setup -- RoBERTa}
\label{sec:roberta_details}
The RoBERTa experiments are based on the Code provided by \citep{attentionGuiding}. We follow the data acquisition and preparation strategy of \cite{megatron}. Thus, we train on the latest Wikipedia dump (downloaded on the  2\textsuperscript{nd} of August 2023). We train a 12 layer RoBERTa model with 12 heads. We use a batch size of 84 and a learning rate of 5e-5.

\subsection{Experimental Setup -- In-Context Learning}
\label{sec:gpt2_details}

The in-context learning experiments are based on the experimental setup of \citet{chan2022data} using the Omniglot dataset \citep{lake2015human}. During training the network is presented with sequences of image label pairs, where each image is followed by it's true label. For the last image the label is missing and must be predicted by the transformer. The sequences are usually constrained in such a way, that the target label is often present at least once in the sequence (burstiness). The task can be partially solved by simply learning to associate image to a label, called in-weights learning (IWL).
To generalize to unseen or rare samples, a better strategy is to exploit the solution given by the example in the context, referred to as in-context learning (ICL).

Experiments are based on the code of \citet{chan2022data}. We follow the original setup and use GPT-2 \citep{radford2019language} with 12 layers, embedding dimension of 64 and 8 heads. The images were embedded using a non pretrained ResNet-18 architecture with (16, 32, 32, 64) channels per group while the labels were embedded using a standard embedding layer.
Following the original training procedure, we ran the experiments for 500k iterations on a single GPU with batch size 32 using the Adam optimizer. We use a learning rate scheduler with a linear Warmup over 4000 iterations to maximum learning rate followed by square root decay. 

We optimized the learning rate for each method independently by testing 5 learning rates in the interval [3e-5, 3e-3]. For each learning rate and method we trained two models with different random seed. For both, the \normsoftmax variant of GPT-2 and GPT-2 a learning rate of 9e-4 lead to best results and was used for the experiments reported here. We ran each experiment with 4 different seeds - 0, 42, 1337 and 80085. We report the averaged results and show 2 instructive examples for each method.

The in-context performance is reported on the 10 holdout classes not seen during training in the 2-way 4-shot few-shot evaluation setting. Here, the sequence consists only of 2 labels (0/1) with 4 images from each and we consider zero in-context performance for random chance level of 50 \%.
For all experiments we used 50\% of burstiness in data and uniform sampling. 

\end{document}